%% file: egbib.tex
\documentclass[10pt,twocolumn,letterpaper]{article}
\usepackage[pagenumbers]{cvpr} 

\usepackage{setspace}

\usepackage[dvipsnames]{xcolor}
\usepackage{mathtools}
\usepackage{listings}
\usepackage{float}
\usepackage{cvpr}
\usepackage{times}
\usepackage{epsfig}
\usepackage{graphicx}
\usepackage{amsmath}
\usepackage{amssymb}
\usepackage{booktabs}
\usepackage{bm}
\usepackage{float}
\usepackage{algorithm}
\usepackage{algorithmic}
\usepackage{multicol}
\usepackage{multirow}
\usepackage{caption}
\usepackage{graphicx}
\usepackage{adjustbox}

\usepackage[pagebackref,breaklinks,colorlinks]{hyperref}

\usepackage[capitalize]{cleveref}
\crefname{section}{Sec.}{Secs.}
\Crefname{section}{Section}{Sections}
\Crefname{table}{Table}{Tables}
\crefname{table}{Tab.}{Tabs.}

\def\cvprPaperID{48} 
\def\confName{CVPR}
\def\confYear{2022}

\usepackage{color}
\definecolor{light}{rgb}{0.5, 0.5, 0.5}

\begin{document}
 
\title{A Closer Look at Few-shot Image Generation}
\author{
Yunqing Zhao$^{1}$
\hspace{1 cm}
Henghui Ding$^{2}$
\hspace{1 cm}
Houjing Huang$^{3}$
\hspace{1 cm}
Ngai-Man Cheung$^{1*}$
\and
$^{1}$ Singapore University of Technology and Design (SUTD)
\hspace{0.8 cm}
$^{2}$ ETH Zürich
\hspace{0.8 cm}
$^{3}$ ByteDance Ltd.
\and
{
\tt\small yunqing\_zhao@mymail.sutd.edu.sg,
}
{
\tt\small ngaiman\_cheung@sutd.edu.sg
}
\vspace{2 pt}
}

\maketitle

\input{Section-0}


\input{Section-1}


\input{Section-2}

\input{Section-3}


\input{Section-4}


\input{Section-5}

\input{Section-6}

\input{Section-7}

\newpage
{\small
\bibliographystyle{ieee_fullname}
\bibliography{egbib}
}

\clearpage
\input{supp}

\end{document}

%% file: Section-0.tex
\begin{abstract}
 Modern GANs excel at generating high quality and diverse images. However, when transferring the pretrained GANs on small target data (\eg, 10-shot), the generator tends to replicate the training samples. 
 Several methods have been proposed to address this few-shot image generation task, but there is a lack of effort to analyze them under a unified framework. 
 As our first contribution, we propose a framework 
 to analyze existing methods during the adaptation.
 Our analysis 
 discovers that while some methods
 have disproportionate focus on diversity preserving which impede quality improvement, all methods achieve similar quality after convergence.
 Therefore, the better methods are those that can slow down diversity degradation.
 Furthermore, 
 our analysis reveals that there is still plenty of room to further slow down diversity degradation.

 Informed by our analysis and to 
 slow down the diversity degradation of the target generator during adaptation, our second contribution proposes to apply mutual information (MI) maximization to retain the source domain's 
 rich multi-level diversity information in the target domain generator.
 We propose to perform MI maximization by contrastive loss (CL), leverage the generator and discriminator as two feature encoders to extract different multi-level features for computing CL.
 We refer to our method as Dual Contrastive Learning (DCL).
 Extensive experiments on several public datasets show that, while leading to a slower diversity-degrading generator during adaptation, our proposed DCL brings visually pleasant quality and state-of-the-art quantitative performance.
 {Project Page:} {\color{RubineRed}\href{https://yunqing-me.github.io/A-Closer-Look-at-FSIG}{yunqing-me.github.io/A-Closer-Look-at-FSIG}}.
\end{abstract}

%% file: Section-1.tex
\section{Introduction}
\label{1}
 Powerful Generative Adversarial Networks (GANs) \cite{goodfellow2014GAN, brock2018bigGAN, karras2020styleganv2} have been built in recent years that can generate images with high fidelity and diversity \cite{wang2020cnn, keshik2021closer}. 
 Unfortunately, these GANs often require large-scale datasets and computational expensive resources to achieve good performance.
 For example, StyleGAN \cite{karras2018styleGANv1} is trained on Flickr-Faces-HQ (FFHQ) \cite{karras2018styleGANv1}, which contains 70,000 images, with almost 56 GPU days.
 When the dataset size is decreased, however, the generator often tends to replicate the training data \cite{feng2021WhenGansReplicate}.

 Would it be possible to generate sufficiently diverse images, given only limited training data? For example, with 10-shot sketch style human faces \cite{wang2008cuhk_sketches}, could we generate diverse face sketch paintings? This \textit{few-shot image generation} task is important in many  real-world applications with limited data, \eg, artistic domains. It can also benefit some downstream tasks, \eg, few-shot image classification \cite{antoniou2017daGAN, chai2021ensembling}.
 
\subsection{A Closer Look at Few-shot Image Generation}
 \label{1.1}
 To address this few-shot image generation task, instead of training from scratch \cite{tseng2021regularizingGAN, yang2021InsGen}, recent literature focus on transfer learning \cite{pan2009yang-qiang-transfer, abuduweili2021semi_transfer, xu2020reliable} based ideas, \ie, leveraging the prior knowledge of a GAN pretrained on a large-scale, diverse dataset of the source domain and adapting it to a small target domain, without access to the source data. 
 The early method is based on  fine-tuning \cite{wang2018transferringGAN}.
 In particular, starting from the pretrained generator $G_s$, the original GAN loss  \cite{goodfellow2014GAN} is used to adapt the generator to the new domain:
\begin{align}
    \label{ladv}
    \min_{G_t}\max_{D_t}\mathcal{L}_{adv} &= \mathbb{E}_{x\sim p_{data}(x)}[\log D_t(x)] \\\notag &+
    \mathbb{E}_{z\sim p_{z}(z)}[\log{(1-D_t(G_t(z)))}],
\end{align}
 where $z$ is sampled from a Gaussian noise distribution $p_z(z)$,  $p_{data}(x)$ is the probability distribution of the real target domain data $x$, $G_t$ and $D_t$ are generator and discriminator of the target domain, and $G_t$ is initialized by the weights of $G_s$. This GAN loss in Eqn. \ref{ladv} forces $G_t$ to capture the statistics of the target domain data, thereby to achieve both good quality (realisticness \wrt target domain data) and diversity, the criteria for a good generator. 

 However, for few-shot setup (\eg only 10 target domain images), such approach is inadequate to achieve diverse target image generation as very limited samples are provided to define $p_{data}(x)$. Recognizing that, recent methods \cite{li2020fig_EWC,ojha2021fig_cdc} have focused disproportionately on improving diversity by preserving diversity of the source generator during the generator adaptation. 
 In \cite{li2020fig_EWC}, Elastic Weight Consolidation (EWC) \cite{kirkpatrick2017ewc_pnas} is proposed to limit changes in some important weights to preserve diversity. In \cite{ojha2021fig_cdc}, an additional Cross-domain Correspondence (CDC) loss is introduced to preserve  the sample-wise distance information of source to maintain diversity, and the whole model is trained via a multi-task loss with the diversity loss $\mathcal{L}_{dist}$ as an auxiliary task to regularize the main GAN task with loss $\mathcal{L}_{adv}$:
\begin{align}
    \label{lcdc}
    \min_{G_t}\max_{D_t} \mathcal{L}_{adv} + \mathcal{L}_{dist}.
\end{align}
 In \cite{ojha2021fig_cdc}, a patch discriminator \cite{isola2017patchGAN, zhu2017cycleGAN} is also used to further improve the performance in $\mathcal{L}_{adv}$. Details of $\mathcal{L}_{dist}$ in \cite{ojha2021fig_cdc}.

 Diversity preserving methods \cite{li2020fig_EWC,ojha2021fig_cdc} have demonstrated impressive results based on Fréchet Inception Distance (FID) \cite{heusel2017FID} which measures the quality and diversity of the generated samples simultaneously. However, second thoughts about these methods reveal some questions:
 \begin{itemize}
    \vspace{-1 mm}
    \setlength{\itemsep}{1pt}
    \setlength{\parsep}{1pt}
    \setlength{\parskip}{1pt}
     \item With disproportionate focus on diversity preserving in recent works \cite{li2020fig_EWC,ojha2021fig_cdc}, will quality of the generated samples be compromised? For example, in 
     Eqn. \ref{lcdc},
     $\mathcal{L}_{adv}$ is responsible for quality improvement during adaptation, but
     $\mathcal{L}_{dist}$ may compete with 
     $\mathcal{L}_{adv}$ as it has been observed in multi-task learning \cite{pmlr-v119-standley20a,Fifty_transference}. We note that this has not been analyzed thoroughly. 
     \item With recent works' strong focus on  diversity preserving \cite{li2020fig_EWC,ojha2021fig_cdc}, will there  still be room to further improve via diversity preserving? How could we know when the gain of diversity preserving approaches become saturated (without excessive trial and error)?
 \end{itemize}

\subsection{Our Contributions}
In this paper, we take the first step to address these research gaps for few-shot image generation. 
Specifically, as our first contribution, we propose to independently analyze the quality and diversity during the adaptation. 
Using this analysis framework, we obtain insightful information on 
quality/diversity progression.
In particular, on one hand, it is true that strong diversity preserving methods such as \cite{ojha2021fig_cdc} indeed impede the progress of quality improvement. 
On the other hand, interestingly, we observe that these methods can still reach high quality rather quickly, and after quality converges they have no worse quality compared to other methods such as 
\cite{wang2018transferringGAN} which uses simple GAN loss (Eqn. \ref{ladv}).
Therefore, methods with disproportionate focus on preserving  diversity  \cite{ojha2021fig_cdc} stand out from the rests as they can produce slow diversity-degrading generators, maintaining good diversity of generated images when their quality reaches the convergence. Furthermore, our analysis reveals that there is still plenty of room to further slow down the diversity degradation across several source $\rightarrow$ target domain setups.

Informed by our analysis, our second contribution is to propose a novel strong regularization to take a further step in slowing down the diversity degradation, with the understanding that it is unlikely to compromise quality as observed in our analysis. Our proposed regularization is based on the observation that rich diversity exists in the source images at different semantic levels: diversity in middle levels such as hair style, face shape, and that in high levels such as facial expression (smile, grin, concentration). However, such source diversity can be easily ignored in the images produced by target domain generators.
Therefore, 
to preserve source diversity information, we propose to maximize the mutual information (MI) between the source/target image features originated from the same latent code, via contrastive loss \cite{oord2018CPC} (CL).
To compute CL, we leverage the generator and discriminator as two feature encoders to extract image features at multiple feature scales, such that we can preserve diversity at various levels. By combining the two feature encoders (generator and discriminator), we gain additional feature diversity. We show that our proposed Dual Contrastive Learning (DCL) outperforms the previous work in slowing down the diversity degradation without compromising the image quality on the target domain, and hence achieving the state-of-the-art performance.

%% file: Section-2.tex
\section{Related Work}
\label{2}

 Conventional few-shot learning \cite{finn2017maml, fei2006one, snell2017prototypical} aims at learning a discriminative classifier \cite{snell2017prototypical, sung2018relationnet, guo2020awgim, milad2021revisit, liu2020crnet_fs_seg}.
 In contrast, generative few-shot learning \cite{ojha2021fig_cdc, li2020fig_EWC, xiao2022few, yang2021one-shot-adaptation} often follows a transfer learning \cite{pan2009yang-qiang-transfer} pipeline to adapt a pretrained GAN on a small target domain, without access to the source data. 
 Specifically, our analysis and comparison in this paper mainly focus on the following recent baseline models:
 \begin{itemize}
    \setlength{\itemsep}{1pt}
    \setlength{\parsep}{10pt}
    \setlength{\parskip}{1pt}
     \item Transferring GAN \cite{wang2018transferringGAN} (\textbf{TGAN}): directly fine-tune all parameters of both the generator and the discriminator on the target domain;
     \item Adaptive Data Augmentation (\textbf{ADA}) \cite{karras2020ADA}: apply data augmentation \cite{zhao2020differentiable, tran2021data_aug_gan, zhao2020image} during adaptation which does not leak to the generator;
     \item \textbf{BSA} \cite{noguchi2019BSA}: only update the learnable scale and shift parameters of the generator during adaptation;
     \item \textbf{FreezeD} \cite{mo2020freezeD}: freeze a few high-resolution layers of the discriminator, during the adaptation process;
     \item \textbf{MineGAN} \cite{wang2020minegan}: use additional modules between the noise input and the generator. It aims at matching the target distribution before input to the generator;
     \item Elastic Weight Consolidation (\textbf{EWC}) \cite{li2020fig_EWC}: apply EWC loss to regularize GAN, preventing the important weights from drastic changes during adaptation;
     \item Cross-domain Correspondence (\textbf{CDC}) \cite{ojha2021fig_cdc}: preserve the distance between different instances in the source.
 \end{itemize}
 We firstly take a rigorous study of the above methods under the same setup. Then, motivated by our findings, we aim to address this few-shot image generation task.

 Our proposed method is related to contrastive learning \cite{oord2018CPC, he2020moco, zhang2019consistencyGAN, caron2020unsupervised, grill2020byol, chen2021simsiam}.
 Contrastive learning for unsupervised instance discrimination aims at learning invariant embeddings with \textit{different} transformation functions of the \textit{same} object \cite{he2020moco, misra2020SSL-1, jaiswal2021SSL-survey}, which can benefit the downstream tasks.
 To slow down the diversity degradation during few-shot GAN adaptation, we present a simple and novel method via contrastive loss \cite{wu2018instance-discrimination, oord2018CPC}: we hypothesize that,
 the same noise input could be mapped to  fake images in the source and target domains
 with shared semantic information
 \cite{park2020contrastive_translation}. In experiments, we demonstrate convincing and competitive results of our approach for few-shot GAN adaptation, with sufficiently high quality and diversity.

%% file: Section-3.tex
\begin{figure*}[ht]
    \centering
    \begin{subfigure}[b]{\textwidth}
        \includegraphics[width=\textwidth]{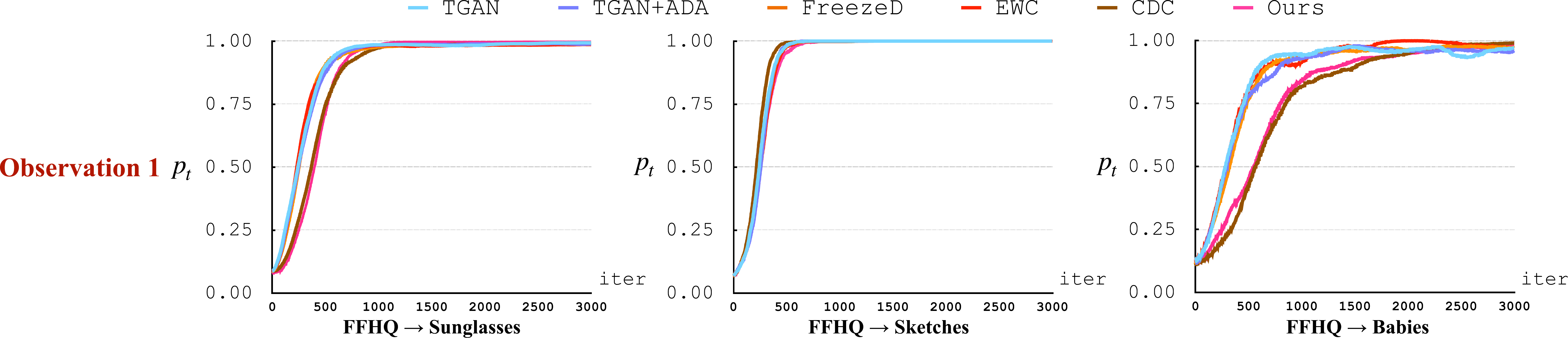}
        \caption{Probability output of the binary classifier (see Sec. \ref{4.2}) \textit{vs.} training iteration during 10-shot adaptation on the target domain.
        }
        \label{2a}
    \end{subfigure}%
    
    \begin{subfigure}[b]{\textwidth}
        \includegraphics[width=\textwidth]{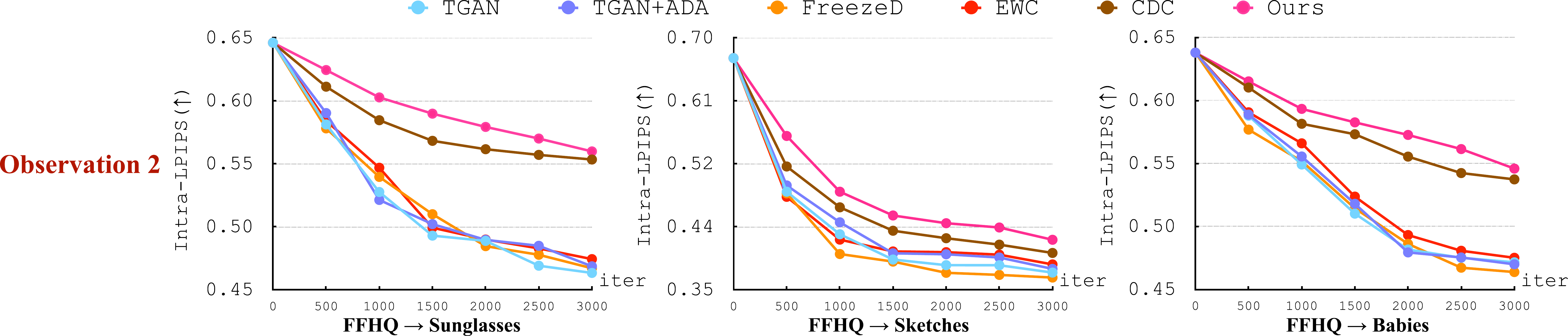}
        \caption{Intra-cluster LPIPS of generated images (see Sec. \ref{4.3}) \textit{vs.} training iteration during 10-shot adaptation on the target domain.
        }
        \label{2b}
    \end{subfigure}%
    \caption{Analysis of existing few-shot GAN adaptation methods under a unified setup (see Sec. \ref{4.4}). 
    \textcolor{BrickRed}{\textbf{Observation 1:}}
    While methods such as CDC
    \cite{ojha2021fig_cdc}
    have disproportionate focus on diversity preserving which impedes quality improvement, all methods we have evaluated achieve 
    \textit{similar} quality/realisticness after convergence.
    \textcolor{BrickRed}{\textbf{Observation 2:}}
    Different methods exhibit substantially \textit{dissimilar} diversity degrading rates.  
    Combined the results in Figure \ref{2a} and \ref{2b}: since the achieved quality is similar, the better methods are those that can slow down diversity degradation.
    We summarize our findings in Sec. \ref{4.5}, which motivate our proposed method in Sec. \ref{5}.
    Best viewed in color.}
    \vspace{-4 mm}
    \label{fig2}
\end{figure*}

\section{Preliminary}
\label{3}

 Given a GAN pretrained on the source domain, we denote the generator as $G_{s}$. In the adaptation stage, the target generator $G_{t}$ and the discriminator $D_{t}$ can be obtained by fine-tuning the pretrained GAN on the target domain via 
 Eqn. \ref{ladv}.
 We follow the settings of the prior work \cite{wang2020minegan, ojha2021fig_cdc, wang2018transferringGAN}: during adaptation, there is \textit{no} access to the abundant source data, and we can only leverage the pretrained GAN and target data for transfer learning. 
 
 For the target data size, some recent work \cite{wang2018transferringGAN, karras2020ADA} applies more than 1,000 images in adaptation, and show satisfied results.  
 In contrast, we focus on the challenging case: fine-tune the pretrained GAN with only few-shot (\eg, 10-shot) data.

%% file: Section-4.tex
\section{Revisit Few-Shot GAN Adaptation}
\label{4}
\subsection{Motivation}
\label{4.1}
 While a few recent works \cite{ojha2021fig_cdc, li2020fig_EWC} propose different ideas, including to inherit the diversity during adaptation, to our knowledge, no effort attends to analyze the underlying mechanism of this problem.
 The specific issues concerning this study are as follows: will quality be compromised due to the disproportionate focus on diversity preserving?
 will there still be room to further improve via diversity preserving?
 We study these concerns from a novel perspective: compare different methods by visualizing and quantifying the few-shot adaptation process. In particular, we decouple their performance by evaluating the quality and diversity on the target domain separately, under a unified framework.
 
\subsection{Binary Classification for Quality Evaluation}
\label{4.2}
 The probability output of a classifier indicates
 the confidence of how likely an input sample belongs to a specific class \cite{hinton-distill, keshik2022ls-kd}. Therefore, we
 employ the probability output of a binary classifier to assess to what extent the generated images belong to the  \textit{target domain}.
 In particular, we train a convolutional network $C$ on two sets of real images (from source and target, excluded during adaptation). Then, we apply $C$ to the synthetic images from the adapted generator $G_t$ during adaptation. The soft output of $C$ are $p_t$, predicted probability of input belonging to the target domain, and ($1-p_t$), that of not belonging to the target domain. Therefore, we take $p_t$ as an assessment of the realisticness of the synthetic images on the target domain: 
 \begin{align}
     p_t &= \mathbb{E}_{z\sim p_{z}(z)}
     [C(G_{t}(z))].
 \end{align}
 $z$ is a batch of random noise input (size=1,000) fixed during adaptation, which is a \textit{proxy} to indicate the quality and realisticness evolution process of different methods.

\subsection{Intra-Cluster Diversity Evaluation}
\label{4.3}
 To evaluate the diversity, we introduce a Learned Perceptual Image Patch Similarity (LPIPS) \cite{zhang2018lpips} based metric, intra-cluster LPIPS (\textit{intra}-LPIPS) \cite{ojha2021fig_cdc}, to identify the similarity between the generated images and the few-shot target samples during adaptation. 
 The \textit{standard} LPIPS
 evaluates the perceptual distance between two images, and it is empirically shown to align with human judgements \cite{zhang2018lpips}. 
 Intra-LPIPS is a variation of the standard LPIPS: We firstly generate abundant (size=1,000) images, then assign each of them to one of the $M$ target samples (with lowest standard LPIPS), and form $M$ clusters. The intra-LPIPS is obtained by computing the average standard LPIPS for random paired images within each cluster, then average over $M$ clusters. We provide the pseudo-code in Supp.
  
 Ideally, we hypothesize that the highest diversity knowledge is achieved by the generator pretrained on the large source domain, and it will degrade during the few-shot adaptation. In the worst case, the adapted generator simply replicates the target images, and intra-LPIPS will be zero. 
 To justify our conjecture, we sample the target generator at different adaptation iterations, then apply intra-LPIPS to assess the diversity of the generated images.

\begin{figure*}[ht]
    \centering
    \includegraphics[width=0.98\textwidth]{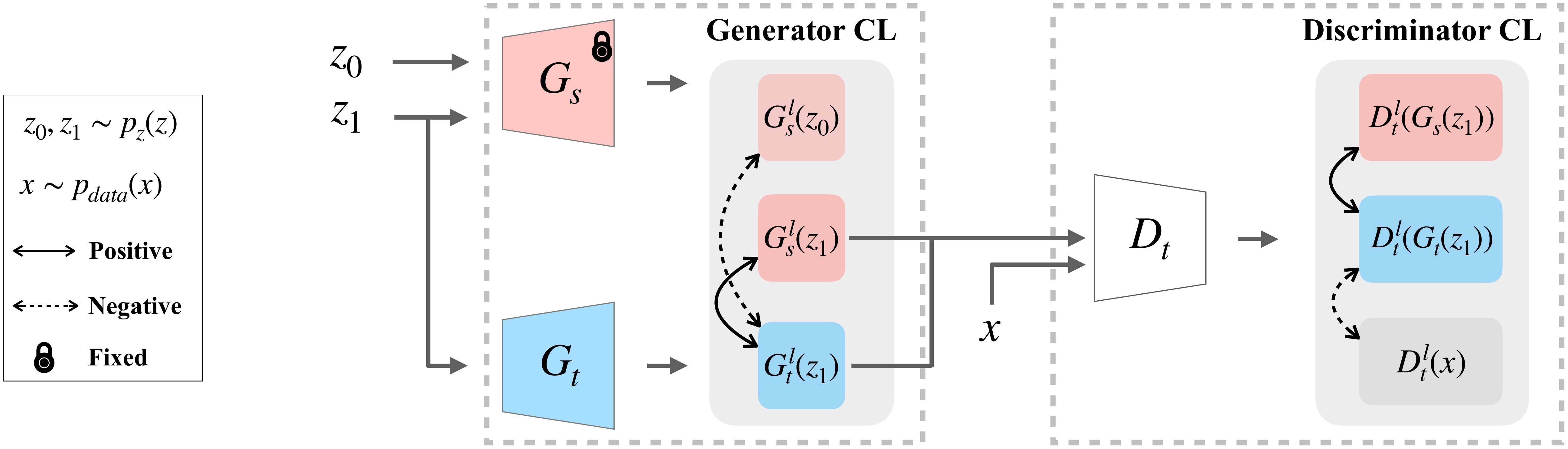}
    \caption{Our proposed Dual Contrastive Learning (DCL) for few-shot GAN adaptation. To slow down diversity degradation during adaptation, in both Generator CL (Eqn. \ref{eq_cl1}) and Discriminator CL (Eqn. \ref{eq_cl2}), we maximize the similarity between positive pairs generated using the same noise input, and push away those that are negative pairs. See details in Sec. \ref{5.1}.} 
    \vspace{-4 mm}
    \label{fig3}
\end{figure*}

\subsection{Experiment settings}
\label{4.4}

 \textbf{Basic setup:} For all methods: We follow the previous work \cite{ojha2021fig_cdc, li2020fig_EWC} to use StyleGAN-V2 \cite{karras2020styleganv2} with default hyperparamters. 
 We use AlexNet \cite{krizhevsky2012alexnet} as the binary classifier.
 We fine-tune the pretrained GAN on the target domain with batch size 4 and 3,000 iterations, on a Tesla V100 GPU.
 
 \textbf{Source $\rightarrow$ target adaptation:}
 Training the binary classifier requires rich data on both source and target domains. Therefore, we transfer the model pretrained on FFHQ \cite{karras2018styleGANv1} (70,000 images) to 10-shot samples from target domains that contain abundant data originally: Sketches, Babies, and Sunglasses (roughly 300, 2500, 2700 images, respectively). 
 To obtain unbiased classifiers, we form each dataset by keeping the source and target training data balanced.
 
 \textbf{Evaluation:} The baseline methods for evaluation are introduced in Sec. \ref{2}. 
 In this section, we do not include BSA \cite{noguchi2019BSA} and MineGAN \cite{wang2020minegan}: BSA applies a different GAN architecture (BigGAN \cite{brock2018bigGAN}), and MineGAN employs additional training modules during adaptation, and it is time-consuming for training. We include the results of these two models in Sec. \ref{6}. To compute the intra-LPIPS, we follow \cite{ojha2021fig_cdc} to use the base implementation from \cite{zhang2018lpips} \protect \footnotemark.
 \footnotetext{\href{https://github.com/richzhang/PerceptualSimilarity}{https://github.com/richzhang/PerceptualSimilarity}}
\subsection{Empirical Analysis Results}
\label{4.5}
 The analysis results are shown in Figure \ref{fig1} and Figure \ref{fig2}. Generally, our observations can be stated as follows:
  \begin{itemize}
    \setlength{\parsep}{1pt}
    \setlength{\parskip}{1pt}
    \item \textbf{Image quality/realisticness} (\textcolor{BrickRed}{\textbf{Observation 1}}): 
    As shown in Figure \ref{2a}, 
    strong diversity preserving methods such as CDC
    \cite{ojha2021fig_cdc} (and ours, to be discussed in Sec. \ref{5})
    indeed  impede  the  progress  of  quality  improvement, most noticeable in FFHQ $\rightarrow$ Babies. This is because additional regularization for diversity preserving (e.g. $\mathcal{L}_{dist}$ in \cite{ojha2021fig_cdc}, see Eqn. \ref{lcdc})  competes with main GAN loss
    $\mathcal{L}_{adv}$ which is responsible for quality improvement. 
    Note that such observation is 
     consistent with findings in multi-task learning \cite{pmlr-v119-standley20a,Fifty_transference}.
    On the other hand, interestingly, we observe that these methods can still reach high quality on the target domain
    rather quickly, and they achieve similar quality as other methods after quality converges.
    We conjecture that, due to the informative prior knowledge on the source domain, it is not difficult to obtain adequate realisticness when transferring to the small target domain.
    \item \textbf{Image diversity}
    (\textcolor{BrickRed}{\textbf{Observation 2}}):
    In Figure \ref{fig1} and Figure \ref{2b}, we demonstrate the qualitative and quantitative results of the diversity change during few-shot adaptation. While different methods can achieve similar quality on the target domain, their diversity-degrading rates vary drastically. For example, in Figure \ref{fig1}, TGAN \cite{wang2018transferringGAN} immediately loses the diversity and the generator replicates the target samples at the early adaptation stage. 
    Since the achieved quality is similar for all these methods, the better methods are those that 
    do not suffer from rapid diversity degradation.
  \end{itemize}
   
  As shown in Figure \ref{2b}, 
  for all existing methods, the loss of diversity is inevitable in adaptation. 
  On the other hand, since  all methods achieve similar realisticness after the quality converges, the better generators are those that can slow down the diversity degradation, and the worse ones are those that lose diversity rapidly before the convergence of the quality improvement.  
  Recalling the concerns we raise in Sec. \ref{4.1}: we show that,
  there is still plenty of room to further reduce the rate of diversity degradation, besides,
  the gain of diversity preservation becomes saturated only until the rate of diversity degradation is much reduced.
  These findings motivate our proposed method in the next section to further slow down  diversity degradation.

%% file: Section-5.tex
\section{Dual Contrastive Learning}
\label{5}

\subsection{Overall Framework}
\label{5.1}
Rich  diversity  exists  in  the  source images $G_s(z)$ at different semantic levels:  
middle levels  such  as  different hair  style, high levels such as different facial expression.
To preserve source images' diversity information in the 
target domain generator, 
we propose to maximize the mutual information (MI) between the source/target image features originated from the same input noise.
In particular, for an input noise 
$z_{i}$, we seek to maximize: 
\begin{align}
    \label{eq_MI}
\text{MI}(\pi^l(G_{t}(z_{i}));\pi^l(G_{s}(z_{i}))),
\end{align}
where $G_{t}(z_{i}), G_{s}(z_{i})$
are generated images by the target / source generators respectively, and $\pi^l(\cdot)$ is a feature encoder to extract $l$-th level features (to be further discussed).
As $G_s$ is fixed during adaptation, the MI maximization enables  $G_t$ to learn to generate images that include source images' diversity at various semantic levels (with the use of multiple $l$ during adaptation).
Furthermore, in GAN, we have  $G_t$ and $D_t$, and we take advantage both of them to use as the feature encoders $\pi^l(\cdot)$ to extract different features.
As directly maximizing MI is challenging
\cite{mutual}, we apply contrastive learning \cite{oord2018CPC} to solve Eqn. \ref{eq_MI}.

Specifically, we propose Dual Contrastive Learning (DCL) for few-shot GAN adaptation, as Figure \ref{fig3}. 
There are several goals for DCL: i) Maximize the MI between generated images on the source/target domain originated from the same noise input; ii) push away the generated images on the source and target domain that use different noise input; iii) push away the generated target images and the real target images, to prevent collapsing to the few-shot target set.
To achieve these goals, we let the generating and the discriminating views using the same noise input, on source and target, as the positive pair, and maximize the agreement between them.
Concretely, DCL includes two parts.
 
 \textbf{Generator CL:} 
 Given a batch of noise input \{$z_{0}$, $z_{1}$, ..., $z_{N-1}$\}, 
 we obtain images
 $\{G_{s}(z_{0}), G_{s}(z_{1}), ..., G_{s}(z_{N-1})\}$ and 
 $\{G_{t}(z_{0}), G_{t}(z_{1}), ..., G_{t}(z_{N-1})\}$, generated on the source and the target domain, respectively. Considering an \textit{anchor} image $G_{t}(z_{i})$, we optimize the following object:
\begin{align}
    \label{eq_cl1}
    \mathcal{L}_{CL_{1}} =
    -\log 
    \frac{f(G^{l}_{t}(z_{i}), G^{l}_{s}(z_{i}))}
    {\sum_{j=0}^{N-1} f(G^{l}_{t}(z_{i}), G^{l}_{s}(z_{j}))}
\end{align}
 which is an \textit{N}-Way categorical cross-entropy loss to classify the positive pair $(G^{l}_{t}(z_{i}), G_{s}^{l}(z_{i}))$ at $l$-th layer correctly. 
 $\frac{
 f(G^{l}_{t}(z_{i}), G^{l}_{s}(z_{i}))}{\sum_{j=0}^{N-1}f(G^{l}_{t}(z_{i}), G^{l}_{s}(z_{j}))}$
 is the prediction probability and 
 \begin{equation}
     \resizebox{.9\hsize}{!}{$
     f(G^l_{t}(z_{i}), G^{l}_{s}(z_{i})) = \exp(CosSim(G_{t}^l(z_{i}), G^{l}_{s}(z_{i}))/\tau)
     $}
 \end{equation}
 for $i \in \{0, 1, ..., N-1\}$ is the exponential of the cosine similarity between $l$-th layer features of the generated images on source and target, scaled by a hyperprameter temperature $\tau=0.07$ and passed as logits \cite{he2020moco, park2020contrastive_translation}. 
 
 \textbf{Discriminator CL:}
  We focus on the view of the discriminator $D_{t}$. Given the generated images by the source and target generator and real data $x$ as input to $D_{t}$, we maximize the agreement between the discriminating features of the generated images on source and target, using the same noise $z_{i}$. To prevent replicating the target data, we regularize the training process by pushing away discriminating features of the generated target images and the real target data at different scales. We optimize the following object:
 \begin{align}
    \label{eq_cl2}
    \mathcal{L}_{CL_{2}} =
    -\log 
    \frac{
    f(D^{l}_{t}(G_{t}(z_{i})), D^{l}_{t}(G_{s}(z_{i})))
    }
    {f(D^{l}_{t}(G_{t}(z_{i})), D^{l}_{t}(G_{s}(z_{i}))) + \Delta
    },
\end{align}
 where $\Delta = \sum_{j=1}^{M} f(D^{l}_{t}(G_{t}(z_{i})), D^{l}_{t}(x_{j}))$
 and $M$ is the number of real target images. The final objective of DCL in our work is simply:
 \begin{equation}
    \min_{G}\max_{D} \mathcal{L}_{adv} + \lambda_{1}\mathcal{L}_{CL{1}} 
     + \lambda_{2}\mathcal{L}_{CL_{2}}.
 \end{equation}
 In practice, we find $\lambda_{1}=2$ and $\lambda_{2}=0.5$ work well. In each iteration, we randomly select different layers of $G_t$ and $D_t$ to perform DCL with multi-level features. 
 
\subsection{Design Choice}
\label{5.2} 
 The design choice of DCL is intuitive: DCL takes the advantage of $G_{s}$=$G_{t}$ before adaptation. Our idea is to re-use the highest diversity knowledge from the source as the strong regularization to slow down the \textit{inevitable} diversity degradation, before the quality improvement converges.
 The Generator CL and Discriminator CL are similar in idea, while they constraint different parts for GAN adaptation: Generator CL requires the adapted generator $G_{t}$ to generate features at different scales that are similar to that of the source generator (with the same noise input). Differently, Discriminator CL regularizes the adversarial training: while fitting to the target domain, the generated target images are encouraged to be pushed away from the real data at different feature scales.

 Under mild assumptions, DCL maximizes the MI (Eqn. \ref{eq_MI}) between the generated samples using the same input noise, on source and target, \eg., for $\mathcal{L}_{CL_1}$: $\text{MI}(G_{t}^l(z_{i});G_{s}^l(z_{i})) \geq \log [N]-\mathcal{L}_{CL_{1}}$ \cite{oord2018CPC}. In the next, we show the effectiveness of DCL in experiments.
 
\setlength{\tabcolsep}{1.5 mm}
\renewcommand{\arraystretch}{1}
\begin{table}[t]  
    \scriptsize
    \centering
    \begin{adjustbox}{width=0.98\columnwidth,center}
        \begin{tabular}{r|c|c|c}
         \multirow{2}{4em} 
         & \textbf{FFHQ $\mapsto$} 
         & \textbf{FFHQ $\mapsto$} 
         & \textbf{FFHQ $\mapsto$} \\
         & \textbf{Babies} 
         & \textbf{Sunglasses} 
         & \textbf{Sketches}\\ 
        \hline
        \textbf{TGAN}\cite{wang2018transferringGAN} & $104.79\pm 0.03$ & $55.61 \pm 0.04$ & $53.42 \pm 0.02$\\
        \textbf{TGAN+ADA}\cite{karras2020ADA} & $102.58 \pm 0.12$ & $53.64 \pm 0.08$ & $66.99 \pm 0.01$\\
        \textbf{BSA}\cite{noguchi2019BSA} & $140.34 \pm 0.01$ & $76.12 \pm 0.01$ & $69.32 \pm 0.02$ \\
        \textbf{FreezeD}\cite{mo2020freezeD} & $110.92 \pm 0.02$ & $51.29 \pm 0.05$ & $46.54 \pm 0.01$\\
        \textbf{MineGAN}\cite{wang2020minegan} & $98.23 \pm 0.03$ & $68.91 \pm 0.03$ & $64.34 \pm 0.02$\\
        \textbf{EWC}\cite{li2020fig_EWC} & $87.41 \pm 0.02$ & $59.73 \pm 0.04$ & $71.25 \pm 0.01$\\
        \textbf{CDC}\cite{ojha2021fig_cdc} & $74.39 \pm 0.03$ & $42.13 \pm 0.04$ & $45.67 \pm 0.02$\\ 
        \textbf{DCL (Ours)} & \bm{$52.56 \pm 0.02$} & \bm{$38.01\pm 0.01$} & \bm{$37.90 \pm 0.02$}\\
        \end{tabular}
    \end{adjustbox}
    \caption{
    For target domains that contain rich data, we follow \cite{ojha2021fig_cdc} to compare FID \protect\footnotemark ($\downarrow$) \cite{heusel2017FID} with different adaptation setups. We firstly generate 5,000 fake images using the adapted generator, then compare them with the real target data (excluded during adaptation). The standard derivations is computed by 5 different runs.
    }
\label{table1}
\end{table}
\footnotetext{\textcolor{black}{Our implementation is based on} \href{https://github.com/mseitzer/pytorch-fid}{https://github.com/mseitzer/pytorch-fid}}

\setlength{\tabcolsep}{1.5 mm}
\renewcommand{\arraystretch}{1}
\begin{table}[t]
    \scriptsize
    \centering
    \begin{adjustbox}{width=0.98\columnwidth,center}
        \begin{tabular}{r | c | c | c }
         \multirow{2}{4em} 
         & \textbf{FFHQ $\mapsto$} 
         & \textbf{Church $\mapsto$} 
         & \textbf{Cars $\mapsto$} \\
         & \textbf{Otto's Paintings} 
         & \textbf{Haunted House} 
         & \textbf{Abandoned Cars} \\
        \hline
        \textbf{TGAN}\cite{wang2018transferringGAN} & $0.51 \pm 0.02 $ & $ 0.52 \pm 0.04$ & $0.46 \pm 0.03$ \\
        \textbf{TGAN+ADA}\cite{karras2020ADA} & $0.54 \pm 0.02$ & $ 0.57 \pm 0.03$ & $0.48 \pm 0.04$\\
        \textbf{BSA}\cite{noguchi2019BSA} & $ 0.46 \pm 0.02 $ & $0.43 \pm 0.02$ & $0.41 \pm 0.03$ \\
        \textbf{FreezeD}\cite{mo2020freezeD} & $ 0.54 \pm 0.03$ & $ 0.45 \pm 0.02$ & $0.50 \pm 0.05$ \\
        \textbf{MineGAN}\cite{wang2020minegan} & $ 0.53 \pm 0.04 $ & $0.44 \pm 0.06$ & $0.49 \pm 0.02$\\
        \textbf{EWC}\cite{li2020fig_EWC} & $0.56 \pm 0.03 $ & $0.58 \pm 0.06$ & $0.43 \pm 0.02 $\\
        \textbf{CDC}\cite{ojha2021fig_cdc} & $ 0.63 \pm 0.03 $ & $0.60 \pm 0.04$ & $0.52 \pm 0.04$ \\
        \textbf{DCL (Ours)} & $\bm{0.66 \pm 0.02}$ & $\bm{0.63 \pm 0.01}$ & $\bm{0.53 \pm 0.02}$\\
        \end{tabular}
    \end{adjustbox}
    \caption{For target domains containing only 10-shot data, we evaluate the diversity of the adapted generator. We firstly generate 5,000 fake images, then we compute intra-cluster LPIPS ($\uparrow$) with few-shot target samples. The standard derivation is computed across 10-shot clusters (see details in Sec. \ref{4.3}).}
    \label{table2}
    \vspace{-3 mm}
\end{table}

%% file: Section-6.tex
 \begin{figure*}[t]
     \centering
     \includegraphics[width=0.985\textwidth]{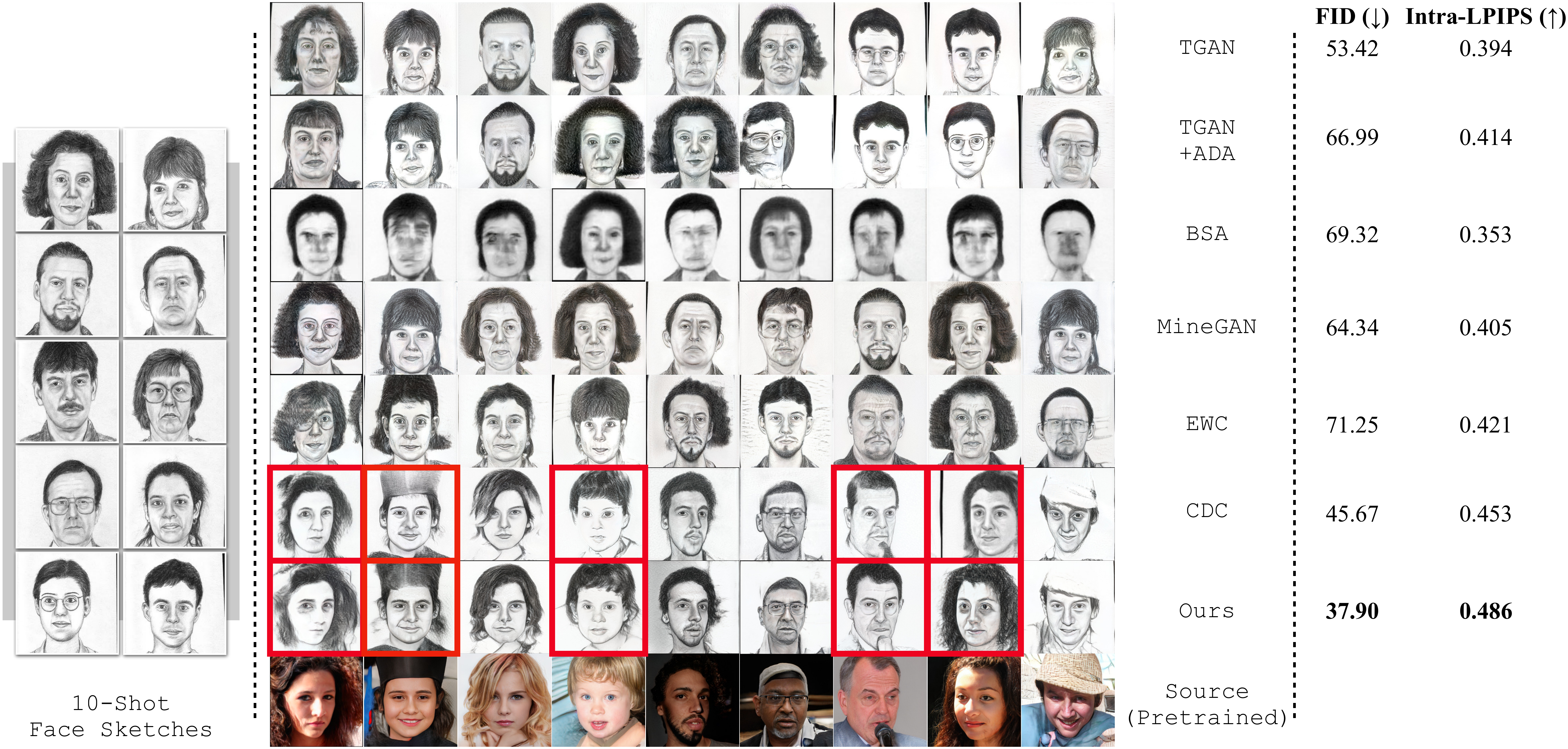}
     \hspace*{0.88 mm}\includegraphics[width=0.985\textwidth]{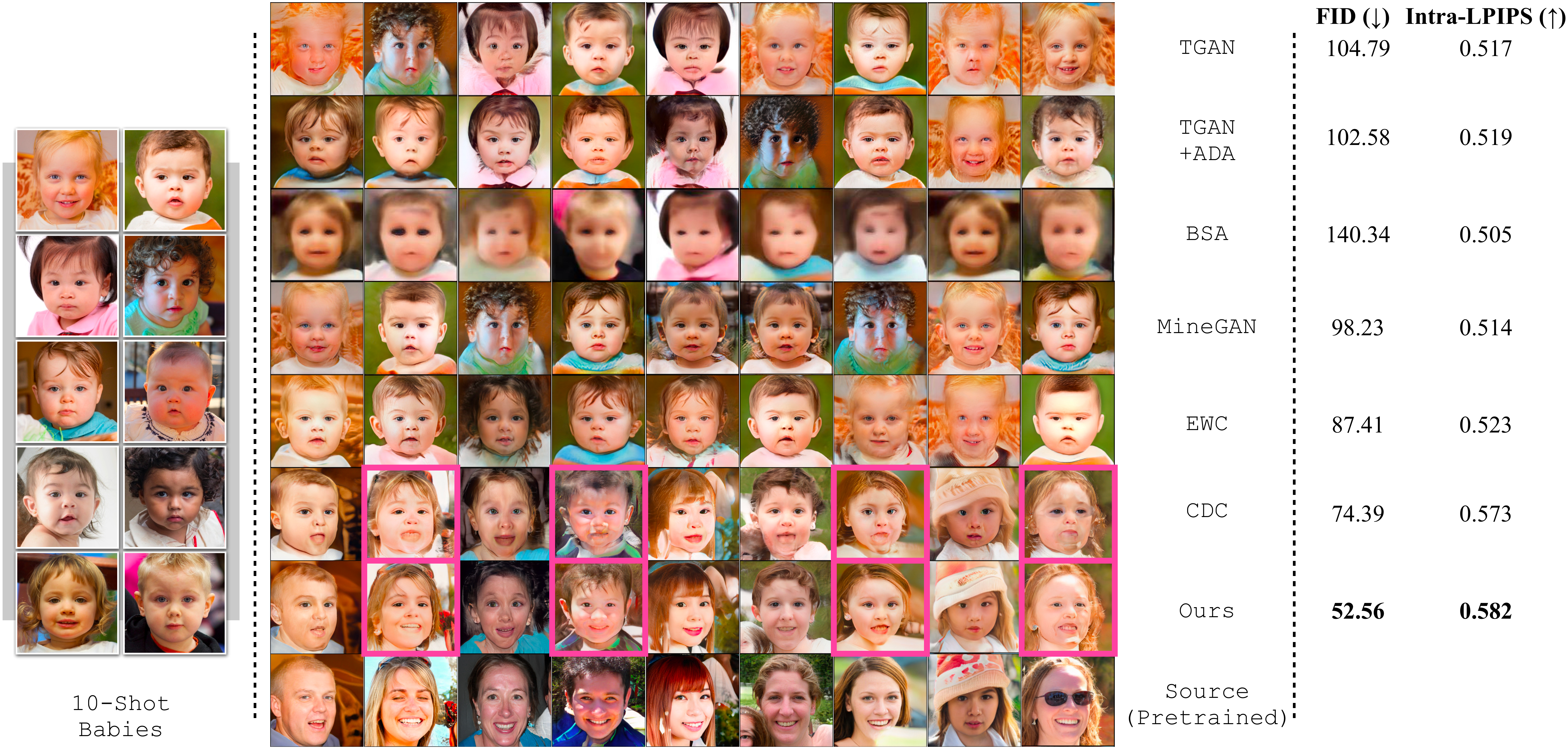}
     \vspace{-3 mm}
     \caption{\textbf{Left}: Transferring a GAN pretrained on FFHQ to 10-shot target samples . We fix the noise input by each column to observe the relationship of generated images before and after adaptation. \textbf{Mid}: We observe that, most of existing methods lose diversity quickly before the quality improvement converges, and tend to replicate the training data. Our method, in contrast, slows down the loss of diversity and preserves more details. For example: In \textcolor{red}{\textbf{red}} frames (\textbf{Upper}), the hair style and hat are better preserved. In \textcolor{Magenta}{\textbf{pink}} frames (\textbf{Bottom}), the smile teeth are well inherited from the source domain. We also outperform others in quantitative evaluation (\textbf{Right}). See details in Sec. \ref{6.2}. 
     }
     \label{fig4}
    \vspace{-6 mm}
 \end{figure*}
 
 \begin{figure*}[t]
     \centering
     \includegraphics[width=0.99\textwidth]{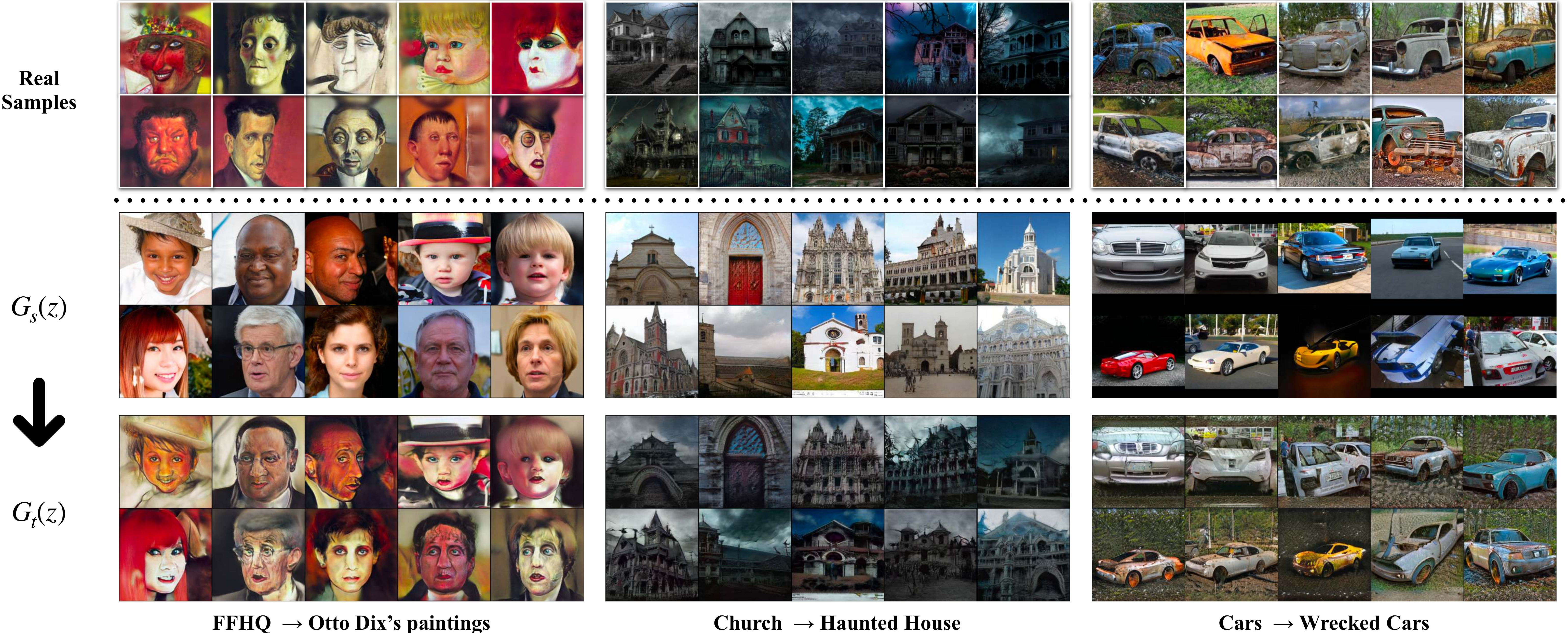}
     \vspace{-2 mm}
     \caption{Generated images with different source $\rightarrow$ target adaptation setups. We show that, the target generator after adaptation with our proposed method can generate visually pleasant images \textit{without much loss of diversity}. The facial expressions of the characters, the structure appearance of the building, and the form/color of vehicles are well preserved and corresponded. See quantitative comparison in Table \ref{table2}.}
     \label{fig5}
     \vspace{-4 mm}
 \end{figure*}

\section{Experiments}
\label{6}

\subsection{Implementation Details}
\label{6.1}
 In this section, we discuss our main experiments. Additional analysis and discussion are included in Supp.
 
 \textbf{Basic setups:} We mostly follow the experiment setups as Ojha \etal \cite{ojha2021fig_cdc}, including the non-saturating GAN loss $\mathcal{L}_{adv}$ in Eqn. \ref{lcdc}, the same pretrained GAN and 10-shot target  samples, with no access to the source data. We employ StyleGAN-V2 \cite{karras2020styleganv2} as the GAN architecture for pretraining and few-shot adaptation, with an image/patch level discriminator \cite{ojha2021fig_cdc}. We fine-tune with batch size 4 on a Tesla V100 GPU. The baseline methods are introduced in Sec. \ref{2}. 
 
 \textbf{Datasets:}
 We use models pretrained on different source domains: i) FFHQ \cite{karras2018styleGANv1} ii) LSUN Church and iii) LSUN Cars \cite{yu15lsun}. We transfer the pretrained model to the following target domains: i) Sketches \cite{wang2008cuhk_sketches} ii) Face paintings by Amedeo and Otto Dix \cite{yaniv2019faceofart} iii) FFHQ-Babies iv) FFHQ-Sunglasses v) Haunted houses vi) Wrecked cars.
 Images for training and evaluation are interpolated to the resolution of 256 $\times$ 256.

\subsection{Comparison with State-of-the-art Methods}
\label{6.2}
 \textbf{Qualitative results.}
 In Figure \ref{fig4}, we visualize the generated images with different methods after adaptation on few-shot target samples.
 In Figure \ref{fig5}, we show more results with different source $\rightarrow$ target adaptation setups.
 Regularized by DCL, the adapted target generator needs to retain the connection to the generated images on source, a slow diversity degradation is thus achieved. We show that, the refined details, \eg, the smile teeth, hairline or structure appearance are well preserved, while the style and texture features are fitted to the target domain. Compared with recent state-of-the-art methods, these semantic meaningful features in the source domain are well inherited by DCL.
 
 \textbf{Quantitative comparison.}
 We mainly focus on two metrics to evaluate our method.  i) For datasets which contain a lot of real data, \eg, FFHQ-Babies, FFHQ-Sunglasses, we apply the widely used Fr\'echet Inception Distance (FID) \cite{heusel2017FID} to evaluate the generated fake images. ii) For those datasets which contain only few-shot real data, FID becomes tricky and unstable, since it summarizes the quality and diversity to a single score. Therefore, we use intra-LPIPS (see Sec. \ref{4.3}) to measure the diversity of generated images. The better diversity comes with a higher score.
 As shown in Table \ref{table1} and Table \ref{table2}, our method outperforms the baseline methods. This indicates that the target generator we obtain can cover a wide range of modes, and the loss of diversity is further reduced.

 \begin{figure}[t]
     \centering
     \includegraphics[width=0.233\textwidth]{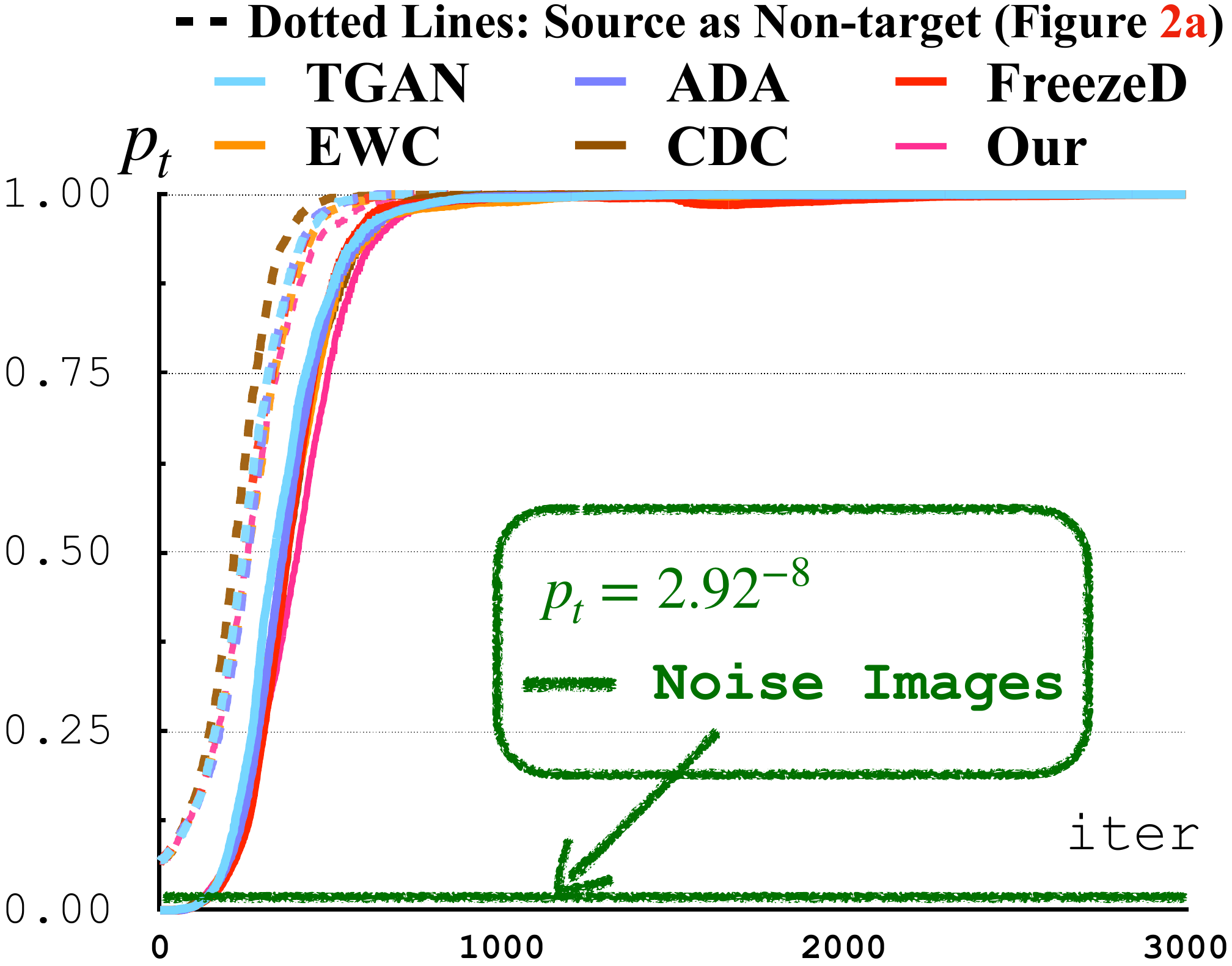}
     \includegraphics[width=0.233\textwidth]{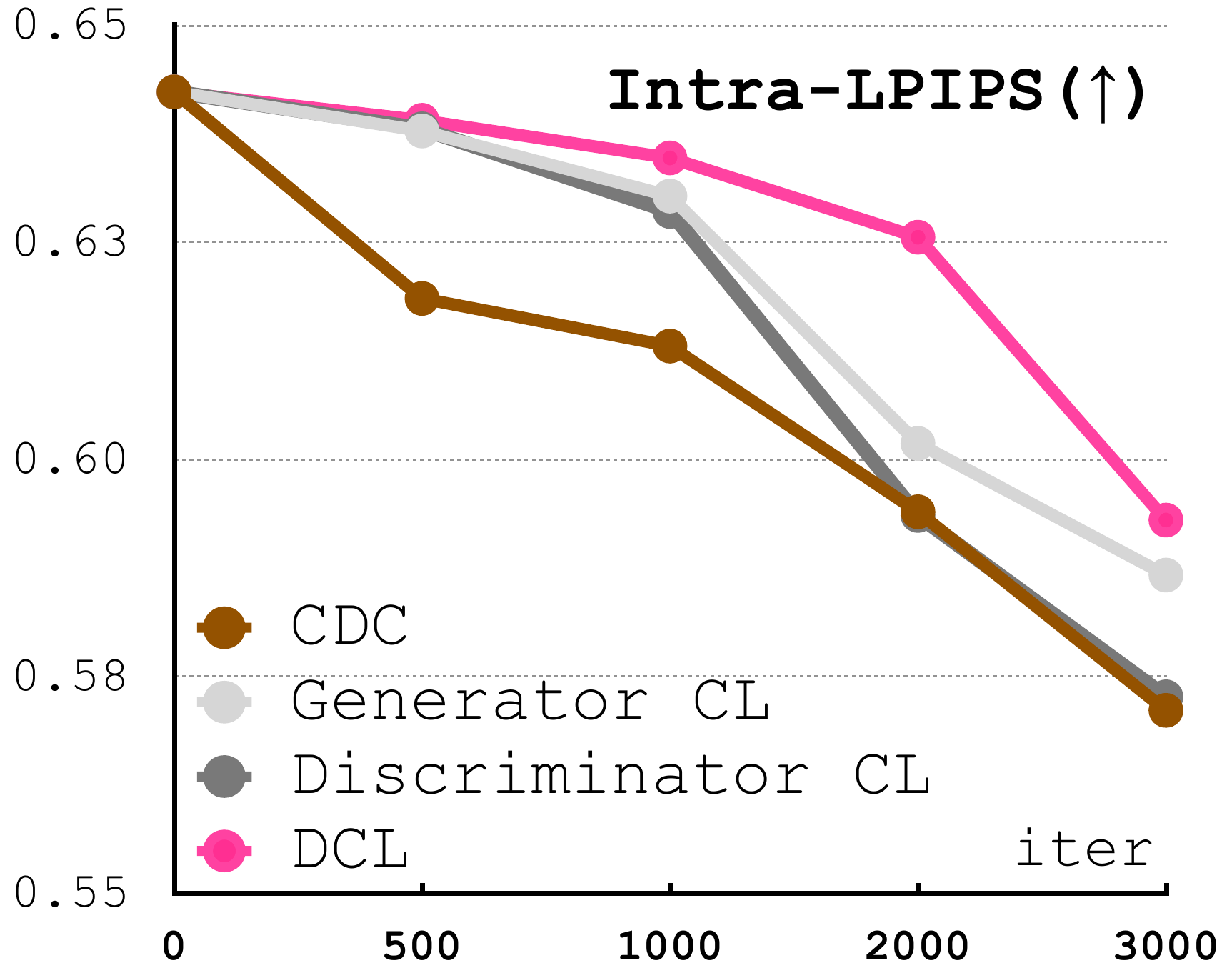}
     \vspace{-1 mm}
     \caption{
     \textbf{Left}: Ablation of the realisticness classifier. We show that 1): Using alternative datasets as non-target to train the classifier give similar results. 2): Moreover, if we input randomly generated noise images, the classifier will give low confidence of being target. 
     \textbf{Right}: Ablation study of DCL, with the same setup as Figure \ref{fig1}.
     }
     \label{fig6}
     \vspace{-4 mm}
 \end{figure}
 
\subsection{Analysis}
\textbf{Effect of our methods.}
 The goal of our proposed binary classifier is to detect if an input image is ``non-target'' or ``target''. In Figure \ref{fig6} (left), we replace the source with equally sampled images from \{FFHQ, ImageNet \cite{deng2009imagenet}, CUB \cite{welinder2010caltech}, Cars \cite{fei-fei-li2013-cars}\} as ``non-target'' and Sketches as ``target'', and we observe the similar results, compared to Figure \ref{2a}.
 We also show that both Generator CL and Discriminator CL can slow down the diversity degradation, and the better results are achieved when combined together. This further confirms our observations in Sec. \ref{4}: the slower diversity degradation leads to the better target generator for few-shot adaptation.
 
\textbf{Effect of target data size.}
In this work, we mainly focus on 10-shot adaptation, similar to the prior literature \cite{ojha2021fig_cdc, li2020fig_EWC}. We further perform ablation study on 5-shot and 1-shot adaptation, use the same setup as Sec. \ref{4} and Sec. \ref{6}. As results in Supp., we have the similar qualitative and quantitative observations, which confirms our findings stated in Sec. \ref{4.5}.

%% file: Section-7.tex
\vspace*{-0.3 mm}
\begin{spacing}{0.972}
\section{Conclusion}
 
 Focusing on few-shot image generation, 
 our first contribution is to analyze 
 existing few-shot image generation methods in a unified framework to gain insights on quality/diversity progress during adaptation.
 Of surprise is that the achieved quality is similar for different existing methods despite some of them have disproportionate focus on preserving diversity.
 Of interest is that the different rates of diversity degradation is the main factor for different performances.
 Informed by our analysis, as our second contribution we propose a mutual information based method to slow down degradation of diversity during adaptation. We connect the source and the target generators through the latent code, and construct positive-negative pairs to facilitate the contrastive learning.
 Our proposed method shows successful results both visually and quantitatively.
 Our study provides concrete information that future work could continue to focus on  reducing the rate of diversity degradation in order to further improve few-shot image generation.
 
 {\bf Acknowledgments:} 
 This research is supported by the National Research Foundation, Singapore under its AI Singapore Programmes (AISG Award No.: AISG2-RP-2021-021; AISG Award No.: AISG-100E2018-005). 
 This project is also supported by SUTD project PIE-SGP-AI-2018-01. 
 We thank anonymous reviewers and Chandrasegaran Keshigeyan for their fruitful discussion and kind help.
 \end{spacing}

%% file: supp.tex
\renewcommand{\thetable}{\color{red}  A\arabic{table}}
\renewcommand{\thefigure}{\color{red} A\arabic{figure}}
\appendix

\section*{Overview of Appendix}
 This  Supplementary  provides  additional  experiments  and  results  to further support our main finding and proposed method for few-shot image generation. The Supplementary materials are organized as follows:
  \begin{itemize}
    \setlength{\itemsep}{1pt}
    \setlength{\parsep}{1pt}
    \setlength{\parskip}{1pt}
     \item Section \textcolor{red}{\ref{A}}: Limitations
     \item Section \textcolor{red}{\ref{B}}: Potential Social/Ethic Impacts
     \item Section \textcolor{red}{\ref{C}}: Additional Details for Binary Classifier
     \item Section \textcolor{red}{\ref{D}}: Pseudo-code for Intra-LPIPS
     \item Section \textcolor{red}{\ref{E}}: Additional Training Details for DCL
     \item Section \textcolor{red}{\ref{F}}: Proof of MI Maximization
     \item Section \textcolor{red}{\ref{G}}: Additional Evaluation Metric
     \item Section \textcolor{red}{\ref{H}}: Additional Results for Source $\mapsto$ Target Adaptation
     \item Section \textcolor{red}{\ref{I}}: Effect of Unrelated Source $\mapsto$ Target Adaptation
     \item Section \textcolor{red}{\ref{J}}: Effect of the Target Data Size 
     \item Section \textcolor{red}{\ref{K}}: Discussion of the Design Choice of DCL
     \item Section \textcolor{red}{\ref{L}}: Does our realisticness classifier learn meaningful attributes?
 \end{itemize}
 
\section{Limitations}
 \label{A}
 We follow exactly previous work (\eg, \cite{ojha2021fig_cdc}) in the choices of domains and datasets for fair comparison. 
 However, given the extremely wide range of domains to which few-shot image generation can be applied, it is not feasible for us to validate our findings for all possible domains.
 On the other hand, our comprehensive qualitative and quantitative experiment results supported by our analysis provide supportive evidence that our findings could be generalized for other domains.
 
 Furthermore, similar to existing work \cite{li2020fig_EWC,ojha2021fig_cdc}, our main focus is on related source/target domains, while we also discuss some analysis on unrelated source/target domains, \eg., see details in Figures \ref{unrelated}. 

\section{Potential Social/Ethic Impacts}
 \label{B}
 In this work, we adhere to the general ethical conducts and guidelines, including that we use the publicly available datasets to conduct all of our experiments, without any personally identifiable information or sensitive identifiable information (\eg, name of the human data). However, since real images are used for transfer learning, we hope the community could take the privacy issue carefully and seriously.
 
 
\section{Additional Details for Binary Classifier}
 \label{C}
 In this section, we provide more details of how to build the binary classifier $C$ (see Sec. \textcolor{red}{4.2} in the main paper) for quality/realisticness evaluation for different methods, during the few-shot adaptation. 
 
 \textbf{Dataset.} As mentioned in the main paper, we aim to build the unbiased binary classifier $C$ by keeping the training data of the source and the target domain balanced. Note that the data used for training the binary classifier is unseen during few-shot adaptation. We summarize the dataset setups and the data link in Table \ref{binary_setup}. 
 
 \textbf{Optimization.} In the training phase of $C$, we randomly initialize the AlexNet with official Pytorch implementation, and we employ the Adam optimizer with binary cross-entropy loss to optimize the weights. We train it until convergence on each dataset. 

  \begin{table}[ht]
    \scriptsize
     \centering
     \vspace{-2 mm}
     \caption{We provide the training data of different source $\mapsto$ target adaptation setups for training the binary classifier $C$.}
     \adjustbox{width=0.47\textwidth}{
     \begin{tabular}{l| c | c | c}
        \toprule
          Source $\mapsto$ Target & Source data & Target data & Size \\\midrule
          FFHQ $\mapsto$ Sketches  & \href{https://drive.google.com/file/d/1NLT_8A4E_Xj3fwxi16CKKgxbRSajAea0/view?usp=share_link}{Link} & \href{https://drive.google.com/file/d/1AJjYdNTaMnnYGCd2k5_WIxjuydJ5RBkK/view?usp=share_link}{Link} & $\sim$ 300 \\
          FFHQ $\mapsto$ Babies  & \href{https://drive.google.com/file/d/1Kx-ei1p1F7CGkpwp_uwdjCFJjf-lKhR6/view?usp=share_link}{Link} & \href{https://drive.google.com/file/d/1zjaF3FzDtLfeYWMrWzktOHiElLcfxKaV/view?usp=share_link}{Link} & $\sim$ 2700\\
          FFHQ $\mapsto$ Sunglasses & \href{https://drive.google.com/file/d/1rib1fNC5WM-zBst7rev8B2MFF5UAZkVc/view?usp=share_link}{Link} & \href{https://drive.google.com/file/d/1RnO7bGvMRx270LSAE3iptoJSicIbDiVd/view?usp=share_link}{Link} & $\sim$ 2500\\
        \bottomrule
     \end{tabular}}
     \label{binary_setup}
     \vspace{-5 mm}
 \end{table}

\section{Pseudo-code for Intra-LPIPS}
 \label{D}
 Intra-LPIPS \cite{ojha2021fig_cdc} evaluates to what extent the generated images collapse to the few-shot target data. The detailed text description of Intra-LPIPS can be found in Sec. \textcolor{red}{4.3} in the main paper.
 In this section, we provide the pseudo-code to compute the intra-LPIPS for evaluating the diversity-degradation during few-shot adaptation, as Algorithm \ref{algo1}. 

\begin{algorithm}[h]
\caption{Pseudo-code of Intra-LPIPS} 
\label{algo1}
\definecolor{codegreen}{rgb}{0,0.6,0}
\definecolor{codegray}{rgb}{0.5,0.5,0.5}
\definecolor{codepurple}{rgb}{0.58,0,0.82}
\definecolor{backcolour}{rgb}{0.95,0.95,0.92}

\lstdefinestyle{mystyle}{
    backgroundcolor=\color{backcolour},   
    commentstyle=\color{codegreen},
    keywordstyle=\color{magenta},
    numberstyle=\tiny\color{codegray},
    stringstyle=\color{codepurple},
    basicstyle=\ttfamily\footnotesize,
    breakatwhitespace=false,         
    breaklines=true,                 
    captionpos=b,                    
    keepspaces=true,                 
    numbers=left,                    
    numbersep=5pt,                  
    showspaces=false,                
    showstringspaces=false,
    showtabs=false,                  
    tabsize=2
}
\lstset{style=mystyle}

\begin{lstlisting}[language=Python]
# Input: 1. Generated images X=[x1, ..., xn];
#        # suppose we have 2-shot target samples;
#        2. Cluster center: c0, c1; 
#        3. Cluster_0, Cluster_1 = [], []
# Output: Avg Intra-LPIPS over 2 clusters
# ------------------------------------------- #
# Step 0. Define the LPIPS function
lpips_fn = lpips.LPIPS(net='vgg')

# Step 1. Assign images to the closet center
for X[i] in X:
    dist0 = lpips_fn(X[i], c0)
    dist1 = lpips_fn(X[i], c1)
    if dist0 < dist1:
        Cluster_0.append(X[i])
    else:
        Cluster_1.append(X[i])

# Step 2. Compute Intra-LPIPS
lpips_dist = []
While not done:
    for img_i, img_j in Cluster_0:
        lpips_dist.append(lpips_fn(img_i, img_j))
    for img_i, img_j in Cluster_1:
        lpips_dist.append(lpips_fn(img_i, img_j))
return lpips_dist.mean()
# ------------------------------------------- #
\end{lstlisting}
\end{algorithm}

\section{Additional Training Details for DCL}
 \label{E}
 We follow the previous work \cite{ojha2021fig_cdc, li2020fig_EWC, mo2020freezeD} to use the architecture of StyleGAN-V2 with pytorch implementation \footnote{\href{https://github.com/rosinality/stylegan2-pytorch}{https://github.com/rosinality/stylegan2-pytorch}}. We use Adam optimizer to optimize the generator and the discriminator, and use the same hyperparameters and settings in \cite{ojha2021fig_cdc}, including the non-saturating loss $\mathcal{L}_{adv}$.
 For the image resolution applied in this work, except for the adaptation setup ``Cars $\rightarrow$ Wrecked cars'', in which we adopts the 512 $\times$ 512 (this is because the GAN pretrained on LSUN Cars adopts the 512 $\times$ 512 image resolution), we use 256 $\times$ 256 for other adaptation setups in both the pretraining and the adaptation stage. We run our experiments (including those in Sec. \textcolor{red}{4}) on a single Tesla V100 GPU.
 
\section{Proof of MI Maximization}
 \label{F}
 Under mild assumptions, our proposed DCL (see Sec. \textcolor{red}{5}) maximizes the lower bound of mutual information (MI) between generated samples with the same noise input, of the source and the target generator, respectively \cite{oord2018CPC}.
 
 In this section, we show the proof of this statement in the main paper.
 We use $\mathcal{L}_{CL_1}$ (with expectation) for example and show that, $\text{MI}(G_{t}(z_{i});G_{s}(z_{i})) \geq \log [N]-\mathcal{L}_{CL_{1}}$, where
 \begin{align}
    \label{cl1}
    \mathcal{L}_{CL_{1}} = \mathbb{E}_{X}
    -\log 
    \frac{f(G_{t}(z_{i}), G_{s}(z_{i}))}
    {\sum_{j=1}^{N} f(G_{t}(z_{i}), G_{s}(z_{j}))}
\end{align}
 To make it concise, in this section we omit the layer index $l$ used in the main paper. 
 We let $X=\{G_{t}(z_{1}), G_{t}(z_{2}), \dots, G_{t}(z_{N})\}$.
 Follow \cite{oord2018CPC}, we write the optimal probability of this objective function (Eqn. \ref{cl1}) as $p(d=i|X, G_{s}(z_{i}))$ where $[d=i]$ indicates that the sample $X_{i}$ is the `positive' sample $G_{t}(z_{i})$ that corresponds to $G_{s}(z_{i})$. The probability of the generated image which is sampled from $p(G_{t}(z_{i})|G_{s}(z_{i}))$, rather than the random generated image distribution, can be shown as follows:
\begin{align}
    p(d=i|X, G_{s}(z_{i})) &=\frac{p(d=i, X|G_{s}(z_{i}))}{\sum_{j=1}^{N}p(d=j, X|G_{s}(z_{i}))} \\&=
    \frac{p(X_{i}|G_{s}(z_{i}))\prod_{k \neq i} p(X_{k})} {\sum_{j=1}^{N}p(X_{i}|G_{s}(z_{i}))\prod_{k \neq j}p(X_{k})} \\&=
    \frac{\frac{p(X_{i}|G_{s}(z_{i}))}{p(X_{i})}}{\sum_{j=1}^{N}\frac{p(X_{j}|G_{s}(z_{i}))}{p(X_{j})}}.
\end{align}
 Therefore, ${f(X_{i}, G_{s}(z_{i}))}$ is proportional to ${\frac{p(X_{i}|G_{s}(z_{i}))}{p(X_{i})}}$. Then, we argue that DCL is a lower bound of the MI between $G_{s}(z_{i})$ from the source generator, and $G_{t}(z_{i})$ from the target generator, which adopt the same noise vector $z_{i}$. This can be shown as follows.
\begin{align}
    \mathcal{L}_{CL_{1}} &= \mathbb{E}_{X}-\log \left\{ \frac{\frac{p(X_{i}|G_{s}(z_{i}))}{p(X_{i})}}
    {\sum_{j=1}^{N}\frac{p(X_{j}|G_{s}(z_{i}))}{p(X_{j})}} \right\}\\&=
    \mathbb{E}_{X}-\log \left
    \{\frac
    {\frac{p(X_{i}|G_{s}(z_{i}))}{p(X_{i})}}
    {\frac{p(X_{i}|G_{s}(z_{i}))}{p(X_{i})} +  \sum_{x\in Neg}\frac{p(X_{j}|G_{s}(z_{i}))}{p(X_{j})}} \right
    \} \\ &=
    \mathbb{E}_{X}\log \left
    \{1+ \frac{p(X_{i})}{p(X_{i}|G_{s}(z_{i}))}
    {\sum_{x\in Neg}\frac{p(X_{j}|G_{s}(z_{i}))}{p(X_{j})}}
    \right
    \} \\&=
    \mathbb{E}_{X}\log \left
    \{1+ \frac{p(X_{i})}{p(X_{i}|G_{s}(z_{i}))}
    {\mathbb{E}\frac{p(X_{j}|G_{s}(z_{i}))}{p(X_{j})}(N-1)}
    \right
    \} \\&=
    \mathbb{E}_{X}\log \left
    \{1+ \frac{p(X_{i})}{p(X_{i}|G_{s}(z_{i}))}
    {(N-1)}
    \right
    \} \\&=
    \mathbb{E}_{X}\log \left
    \{\frac{p(X_{i}|G_{s}(z_{i}))-p(X_{i})}{p(X_{i}|y)}+ N\frac{p(X_{i})}{p(X_{i}|G_{s}(z_{i}))}
    \right
    \} \\& \geq
    \mathbb{E}_{X}\log \left
    \{N\frac{p(X_{i})}{p(X_{i}|G_{s}(z_{i}))}
    \right
    \} \\&=
    \log [N] - MI(G_{t}(z_{i});G_{s}(z_{i}))
\end{align}
 Therefore, we have $MI(G_{s}(z_{i});G_{s}(z_{i})) \geq \log [N]-\mathcal{L}_{CL_{1}}$, which means the Eqn. \ref{cl1} is a lower bound of the mutual information between $G_{s}(z_{i})$ and $G_{t}(z_{i})$.


\section{Additional Evaluation Metric}
 \label{G}
 \textbf{Standard LPIPS.} In the main paper, we use intra-LPIPS \cite{ojha2021fig_cdc} to evaluate the diversity (degradation) of the target generator for different methods. Here, we provide the standard LPIPS ($\uparrow$) results in order to have a comprehensive comparison.  In Table \ref{table_standard_lpips}, we show that, we still outperform other models. Different from intra-LPIPS, the standard LPIPS only evaluate if the generated images are different from each other, and does not evaluate if they collapse to the few-shot training samples, hence we do not include the result of standard LPIPS in the main paper.
\begin{table}[ht]
        \scriptsize
        \centering
        \caption{Standard Pair-wise LPIPS distance ($\uparrow$) of generated fake images. We firstly generate abundant data using the adapted generator on the target domain, then we compute the average perceptual distance between randomly paired images \cite{zhang2018lpips}.}
        \begin{tabular}{r | c | c |  c}
         \multirow{2}{4em} 
         & \textbf{Church $\mapsto$} 
         & \textbf{FFHQ $\mapsto$} 
         & \textbf{FFHQ $\mapsto$} \\
         & \textbf{Haunted house} 
         & \textbf{Amedeo's paintings} 
         & \textbf{Sketches} \\
        \hline
        \textbf{TGAN}\cite{wang2018transferringGAN} & $0.57 \pm 0.06$ & $0.58 \pm 0.12$ & $ 0.44\pm 0.07$\\
        \textbf{TGAN+ADA}\cite{karras2020ADA} & $ 0.60 \pm 0.05$ & $ 0.61 \pm 0.11$ & $ 0.45 \pm 0.08$\\
        \textbf{BSA}\cite{noguchi2019BSA} & $0.47 \pm 0.05$ & $ 0.45 \pm 0.07$ & $ 0.32 \pm 0.05 $ \\
        \textbf{FreezeD}\cite{mo2020freezeD} & $ 0.55\pm 0.08 $ & $ 0.55\pm 0.13$ & $ 0.42 \pm 0.09$\\
        \textbf{MineGAN}\cite{wang2020minegan} & $ 0.56 \pm 0.10$ & $0.59 \pm 0.12$ & $ 0.46\pm 0.09$\\
        \textbf{EWC}\cite{li2020fig_EWC} & $ 0.59\pm 0.06$ & $ 0.60 \pm 0.09$ & $ 0.45 \pm 0.06$\\
        \textbf{CDC}\cite{ojha2021fig_cdc} & $0.61\pm 0.03$ & $ 0.62 \pm 0.06$ & $ 0.47 \pm 0.05$\\ 
        \textbf{DCL (Ours)} & \bm{$0.63 \pm 0.03$} & \bm{$ 0.64 \pm 0.06$} & \bm{$0.51 \pm 0.05$}\\
       \end{tabular}
        \label{table_standard_lpips}
     \hfill
     \label{table3}
\end{table}

 
\section{Additional Results of Source $\mapsto$ Target Adaptation}
 \label{H}
 In this section, we perform additional source $\mapsto$ target adaptation experiments to visualize the effectiveness of our method. In Figure \ref{fig_diff_adaptation}, compared to the source domain images, the generated samples on the target domain preserve rich semantic features (\eg, hair style, hat, building structure) on the source, but capture the style (and accessories) of the few-shot target set, which further confirm our ideas proposed in this work.

\section{Effect of Unrelated Source $\mapsto$ Target Adaptation}
 \label{I}
 \textbf{Background.} In this work, we mainly focus on the few-shot image generation (with GAN adaptation) where the source domain and the target domain are related, similar to all existing methods \cite{mo2020freezeD, wang2018transferringGAN, li2020fig_EWC, ojha2021fig_cdc}. However, the case where the source and the target domain are unrelated should be included in the discussion, \eg., transferring from FFHQ (human face) to Haunted Houses. 
 
 \textbf{Experiments.} In this section, we compare with other methods (see related works in the main paper) with the setup that the source domain and the target domain are unrelated. Note that all other settings are identical to Sec. \textcolor{red}{6} in the main paper. In Figure \ref{unrelated}, we adapt two source domains (FFHQ, LSUN Church) to three different target domains (Haunted house, Amedeo's paintings and Van Gogh's house). The differences between these methods are more obvious, as discussed in Figure \ref{unrelated}.
 
 Nevertheless, these methods cannot accurately capture the target domain distribution \textbf{with much diversity knowledge}, as what we expect when the source domains and the target domains are related. To the best of our knowledge, there is no existing work that focus on this issue and we leave this open problem as our future work.

\section{Effect of the Target Data Size}
 \label{J}
 In the main paper, we mainly focus on the 10-shot adaptation setups, in both Sec. \textcolor{red}{4} and Sec. \textcolor{red}{6}. Here, we extend our analysis to 5-shot and 1-shot setups. As Figure \ref{1-shot-adaptation} and Figure \ref{5-shot-adaptation}, we show that, our main analysis is still hold for 1-shot and 5-shot adaptation case:
 while some methods have disproportionate focus on diversity preserving which impede quality improvement, they will achieve almost the identical realisticness on the target domain, \textbf{even for 1-shot and 5-shot adaptation}. Therefore, we argue that the main focus of few-shot image generation method should be on reducing the diversity degradation during few-shot adaptation.
 
\section{Discussion of the Design Choice of DCL}
 \label{K}

 \textbf{Choice of coefficients.} In Eqn. \textcolor{red}{8} in the main paper, there are two coefficients: $\lambda_1$ and $\lambda_2$ in the loss term of DCL. We perform a grid search to tune these hyperparameters, depending on the performance on diversity and FID score. In experiments, we empirically find that the setting $\lambda_1 = 2$, $\lambda_2 =0.5$ achieves the best result.
 
 \textbf{Choice of batch size.} For our proposed DCL, in Generator CL, the batch size depends on how many noise vectors we sample in each iteration. In Discriminator CL, the batch size depends on how many real samples we have in the few-shot adaptation. Therefore, for fair comparison, we sample 4 noise vectors as input in each iteration, which is identical to other methods, while we sample all few-shot real target images (\eg., 10-shot) to perform Discriminator CL.
 
\begin{figure}[ht]
 \centering
 \includegraphics[width=0.4\textwidth]{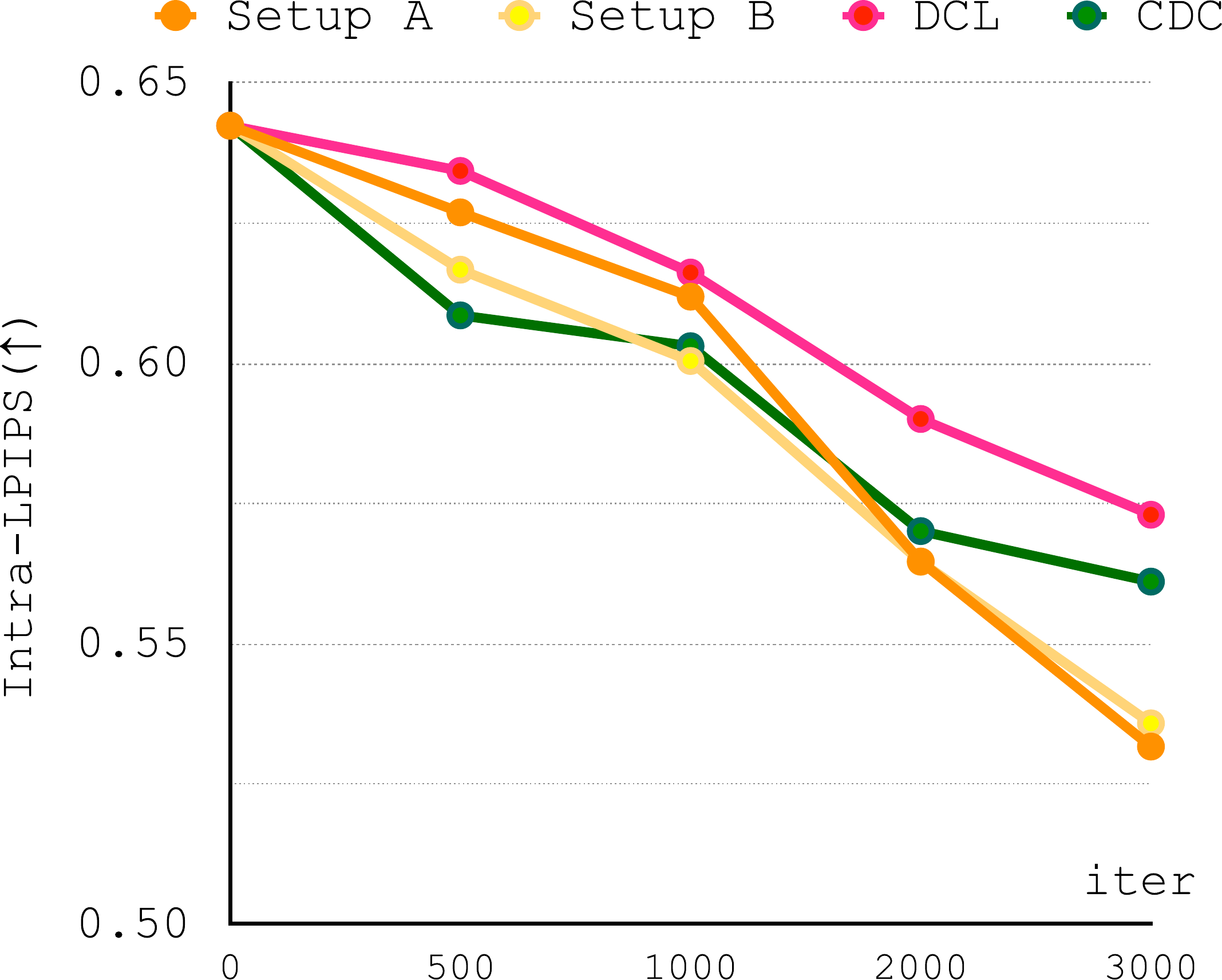}
 \caption{Transferring from FFHQ $\mapsto$ 10-shot Amedeo's paintings (the same setup as Figure \textcolor{red}{1} in the main paper). We show that both Setup A and Setup B have similar performance on mitigating the loss of diversity during few-shot adaptation. Note that we do not use Discriminator CL in this ablation study.}
 \label{choice_neg_ab}
\end{figure}

\begin{figure*}[ht]
 \centering
 \includegraphics[width=0.99\textwidth]{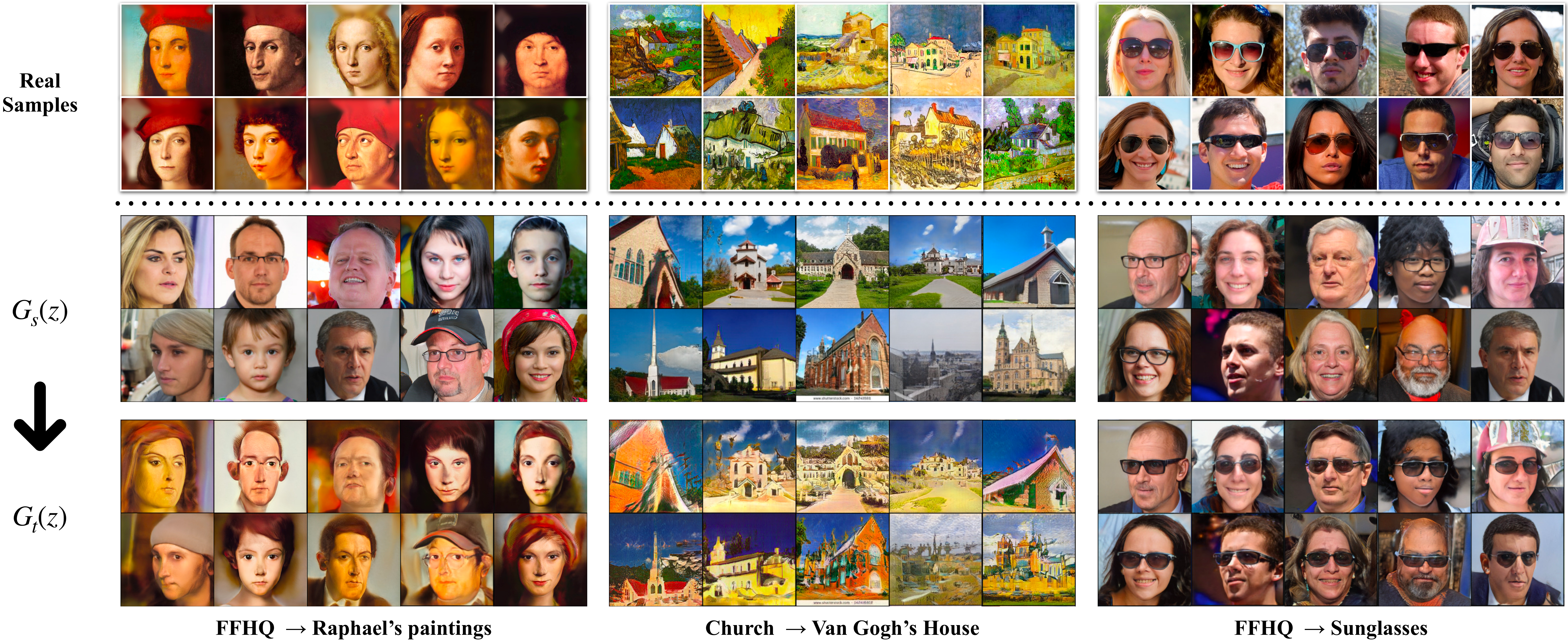}
 \caption{Generated images with additional source $\rightarrow$ target adaptation setups, which is an extension to the Figure \textcolor{red}{5} in the main paper. We show that, while preserving the meaningful semantic features on the source domain (high level: rough structures, face appearance, and huamn postures, middle level: hair style, hat and color), the 
 generated images after adaptation are able to capture good style or texture information on the target domain. }
 \label{fig_diff_adaptation}
\end{figure*}

\begin{figure*}[ht]
     \centering
     \includegraphics[width=0.99\textwidth]{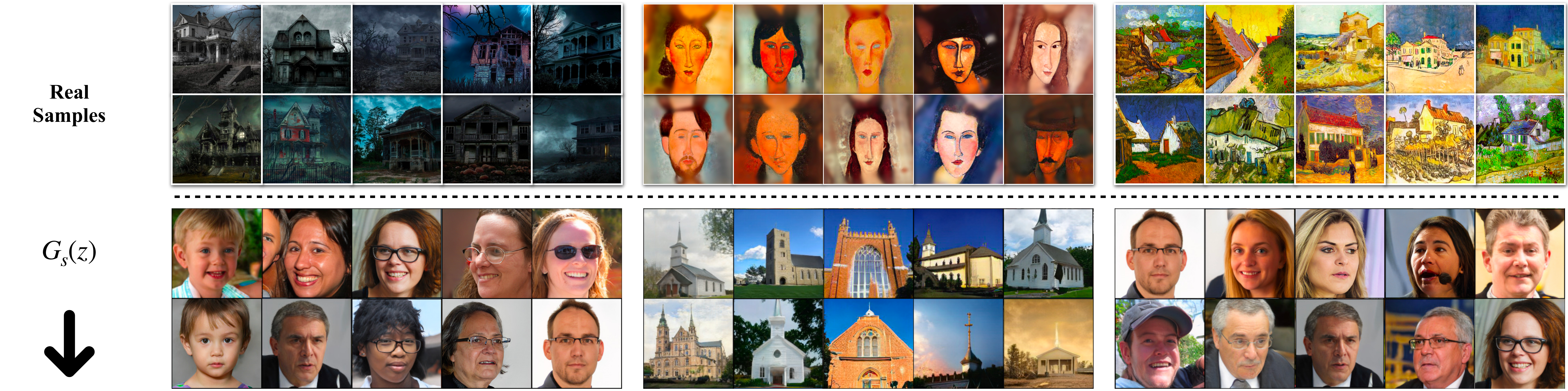}
     \includegraphics[width=0.99\textwidth]{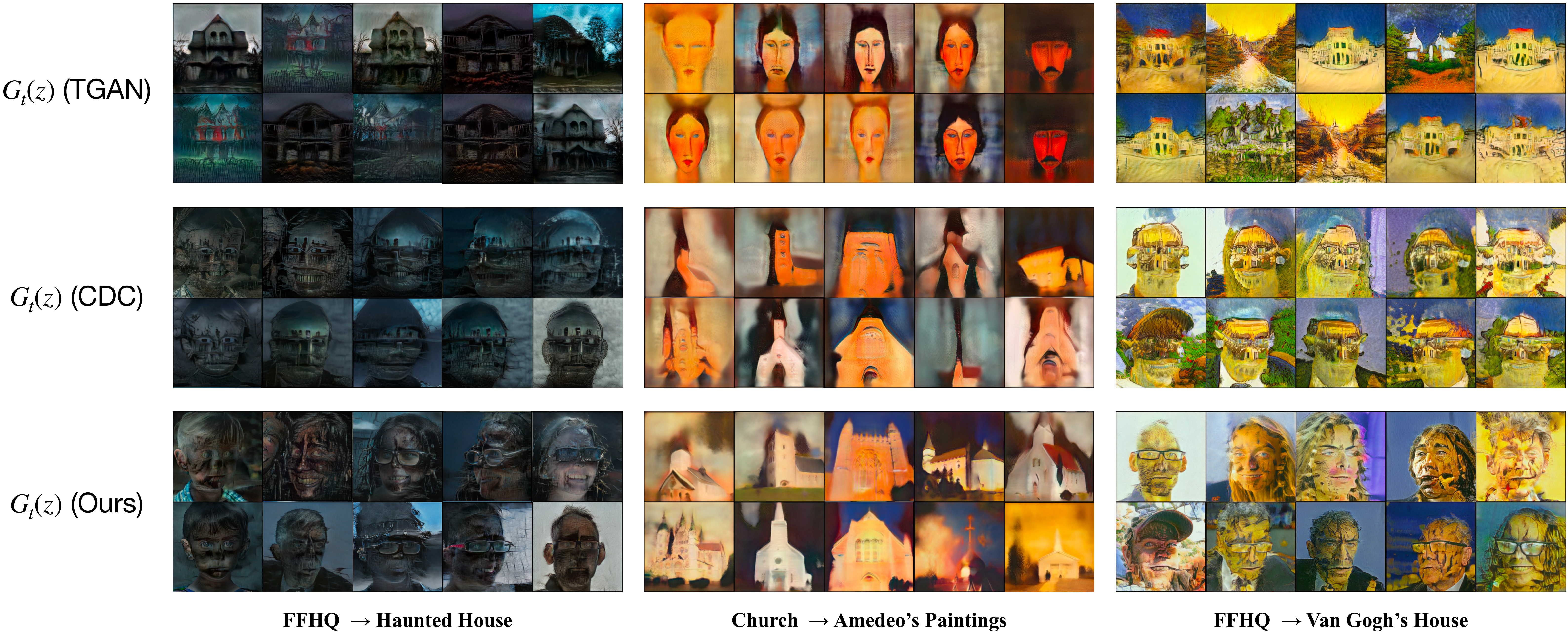}
     \caption{Generated images with \textbf{unrelated} source $\rightarrow$ target adaptation setups. We show that, TGAN still overfits the few-shot target set regardless of the source domain knowledge; CDC preserves the distance between instances in the source, therefore it captures the part-level correspondence between the source and the target, \eg., the eyes and the teeth are roughly mapped to the doors and windows in the target domain. In contrast, since DCL (Ours) emphasizes on the connection to the generated image on the source domain with the same latent code, our results preserve more refined details (\eg, glasses, hair style) and the structure appearance is not thoroughly destroyed when transferring to the target domain. 
     }
\label{unrelated}
 \end{figure*}

\begin{figure*}[ht]
 \centering
    \begin{subfigure}[b]{\textwidth}
        \includegraphics[width=\textwidth]{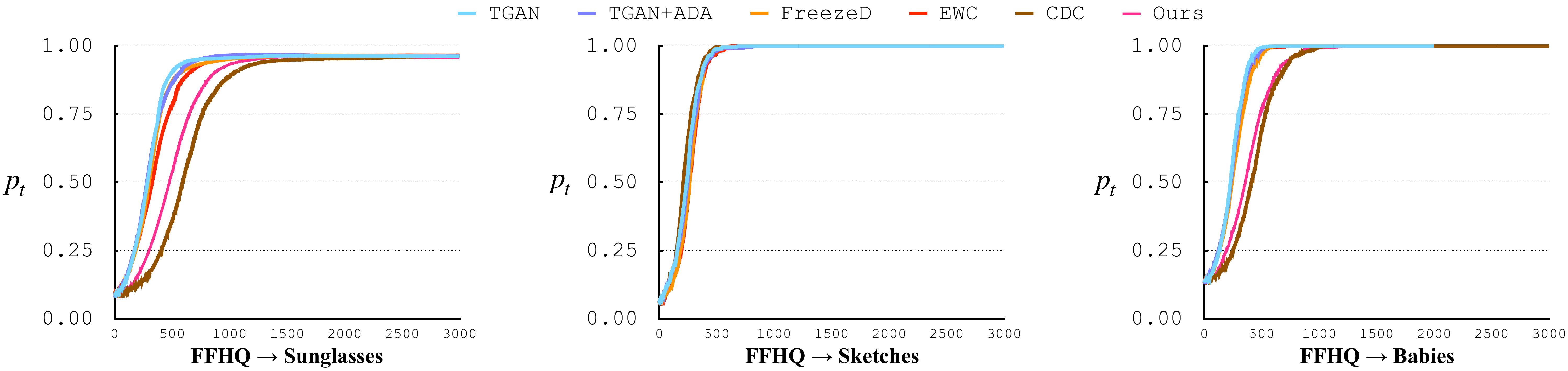}
        \caption{1-shot quality/realisticness evaluation.
        }
        \label{1-shot-quality}
    \end{subfigure}%
    
    \begin{subfigure}[b]{\textwidth}
        \includegraphics[width=\textwidth]{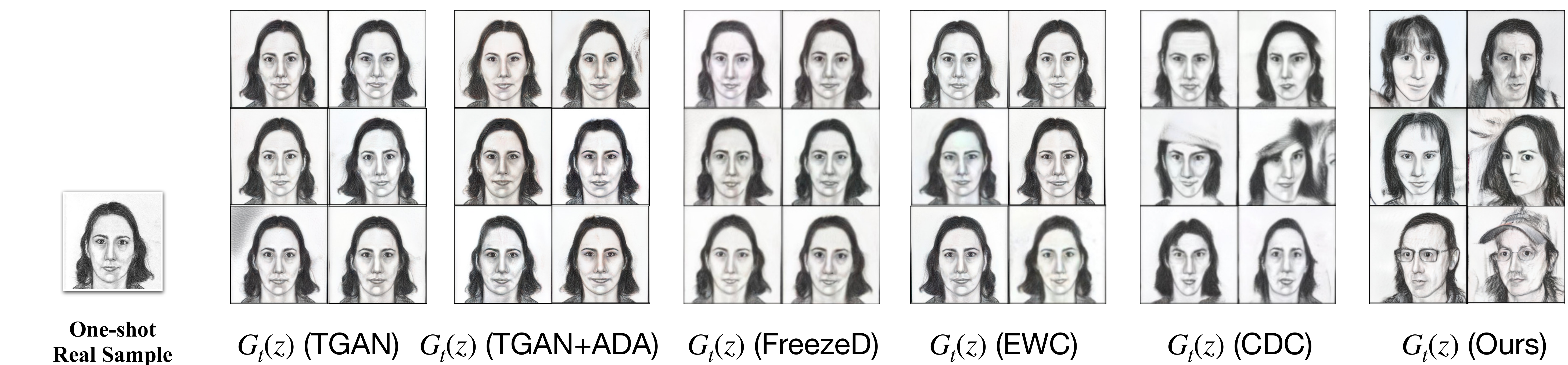}
        \caption{1-shot visualization from FFHQ $\mapsto$ Sketches.
        }
        \label{1-shot-visualization}
    \end{subfigure}%
 \caption{Generated images with 1-shot adaptation (the same setup as Section \textcolor{red}{4} in the main paper), which is an extension to the Figure \textcolor{red}{2} in the main paper.}
 \label{1-shot-adaptation}
\end{figure*}

\begin{figure*}[ht]
 \centering
    \begin{subfigure}[b]{\textwidth}
        \includegraphics[width=\textwidth]{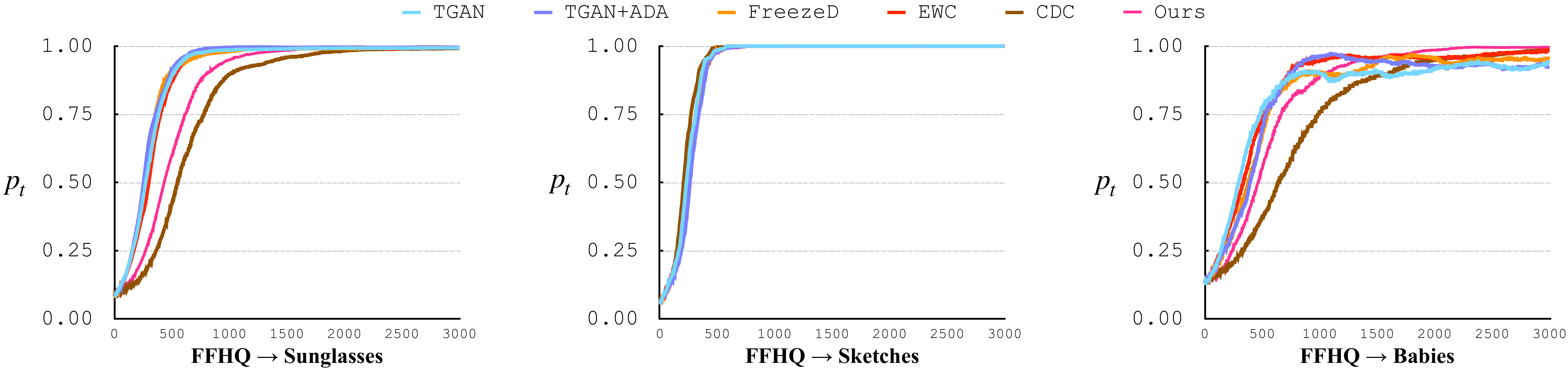}
        \caption{5-shot quality/realisticness evaluation.
        }
        \label{5-shot-quality}
    \end{subfigure}%
    
    \begin{subfigure}[b]{\textwidth}
        \includegraphics[width=\textwidth]{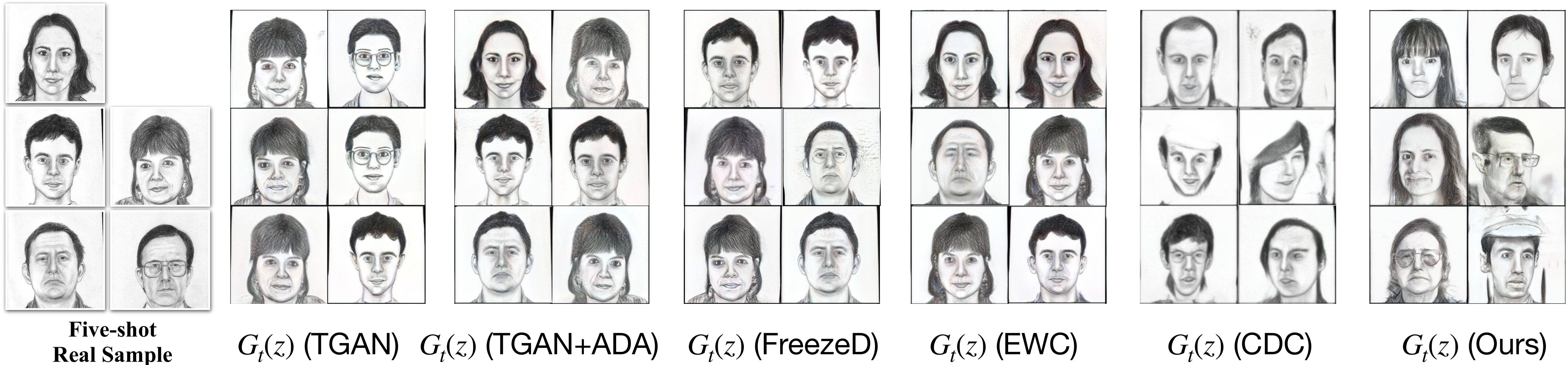}
        \caption{5-shot visualization from FFHQ $\mapsto$ Sketches.
        }
        \label{5-shot-visualization}
    \end{subfigure}%
 \caption{Generated images with 5-shot adaptation (the same setup as Section \textcolor{red}{4} in the main paper), which is an extension to the Figure \textcolor{red}{2} in the main paper.}
 \label{5-shot-adaptation}
\end{figure*}

 \textbf{Choice of negative samples.} The negative samples can be selected from various sources for both Generator CL and Discriminator CL. In Generator CL, the negative samples are $G_{s}(z_{j \neq i})$ where $z_i$ is used to produce $G_{z_{i}}$ (\textbf{Setup A}). However, the negative samples can also be $G_{t}(z_{j \neq i})$ to prevent all generated images of the adapted generator collapsing to the same mode (\textbf{Setup B}). Empirically, we find that the both setups has similar performance on reducing the loss of diversity during adaptation, as we show the change of intra-LPIPS in Figure \ref{choice_neg_ab}. 
 
 For Discriminator CL, we aim to prevent the generated images collapsing to real target data, as observed in other methods (\eg., TGAN). Therefore, we sample discriminating features from real target data (\ie., $D_{t}(x)$) as negative samples to regularize the few-shot adaptation process. Potentially, the negative samples can also come from the generated images. However, in experiments we do not observe better performance with this setup.


\section{Does our realisticness classifier learn meaningful attributes?}
\label{L}

\begin{figure*}[ht]
 \centering
 \includegraphics[width=0.99\textwidth]{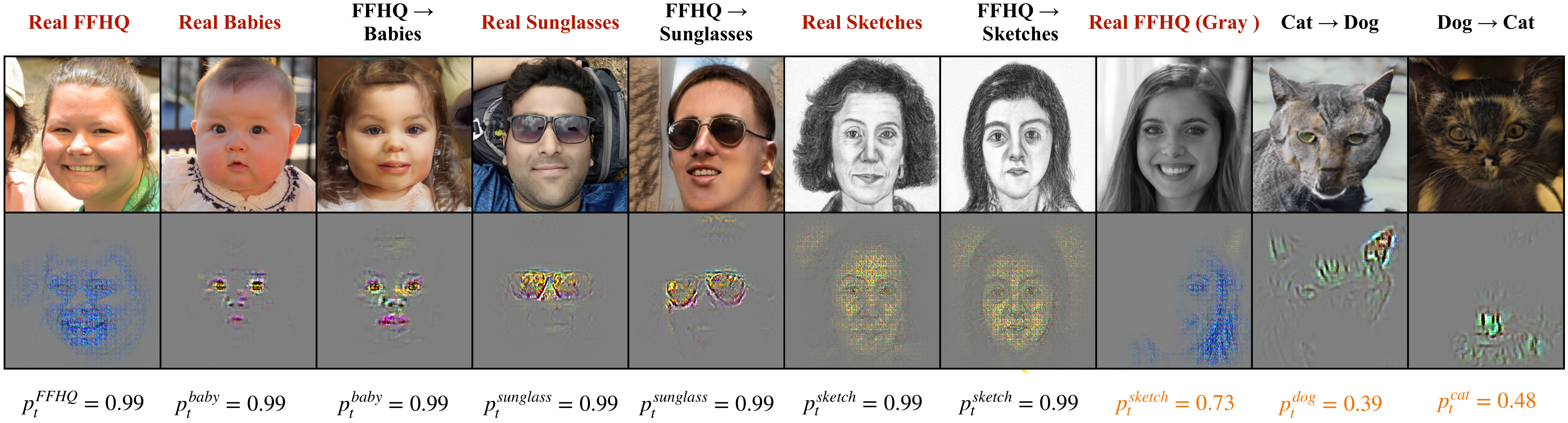}
 \caption{
    We use {GGC} \cite{selvaraju2017grad}   
    to show pixel-wise explanations that our $C$ learns  
    \textit{meaningful semantic features} (not just color)
    to discriminate between source and target domains, 
    The output probability of $C$ for each sample is indicated.
    Pixel-wise explanations clearly show that $C$ leverages on substantial amounts of \textit{meaningful semantic features}
    in discriminating source and target domains:
    $\bullet$ FFHQ: {jawline/lips};
    $\bullet$ FFHQ $\xrightarrow{}$ Babies: {eyes/lips}
    $\bullet$ FFHQ $\xrightarrow{}$ Sunglasses: {sunglass /style}
    $\bullet$ FFHQ $\xrightarrow{}$ Sketches: {style (face)}
    $\bullet$ Dogs $\xrightarrow{}$ Cats: {snouts/ears in dogs, whiskers / fur in cats}.
    $p_t$ are output probabilities of $C$.
    For Dogs $\leftrightarrow$ Cats adaptation, we intentionally select bad images at initial iterations (500) of the adaptation, to validate low $p_t$ from $C$. 
    For other adaptations, images at the end are selected. 
    \textbf{Best viewed in color and enlarged.}
 }
 \label{fig_ggc}
\end{figure*}

We applied Guided Grad-CAM ({GGC}) \cite{selvaraju2017grad} to show pixel-wise explanations and validated that our classifier $C$ (as Sec. {\color{red} 4} in the main paper) indeed learns meaningful semantic features to evaluate the realisticness of generated images, w.r.t. target domain. 
The results are in Figure \ref{fig_ggc}. Additional analysis is below:

\begin{itemize}
    \item 
    We convert real FFHQ images to grey-scale, our $C$ outputs 0.73 of being sketch for 1k images (column 8). The score is not high. We hypothesize that our $C$ learns other features beyond color, \eg style.  
    In this case, {GGC}'s visualization shows face regions are relevant for decision, which is consistent for color/style features. 
    \item
    Importantly, we have visually inspected generated sketch images which have high score from $C$, and they resemble real sketch images (column 7). 
    Overall, we believe our $C$ learns other features beyond color, but it remains open problem to validate what DNN has learned. 
    On the other hand, we notice the quick convergence for FFHQ-to-sketch (as in Figure \ref{fig2} in the main paper). However, it is possible that this is due to easy adaptation: generator only needs to adapt color/style rather than semantic concepts.
    \item 
    The goals of our $C$ and discriminator are similar, but {\em discriminator is trained with very limited target (``real'') samples, preventing it to be used for reliable evaluation}. 
    Meanwhile, $C$ is trained with more abundant held-out target samples. Moreover, as our $C$ is trained with {\em real} target samples, $C$ should evaluate later part of few-shot adaptation as well.
\end{itemize}

Overall, with additional analysis and visualization, we are more confident in our realisticness evaluation in Sec. {\color{red} 4}. The results are consistent and supportive for our main take-away: {\em ``...all methods achieve similar quality after convergence. Therefore, the better methods are those that can slow down diversity degradation. ...there is still plenty of room to further slow down diversity degradation.''}

